%% file: main.tex
\documentclass{article} %
\usepackage{iclr2026_conference,times}

\input{headers/math_commands.tex}

\input{headers/header-lqa}

\input{headers/header}

\input{headers/header-lqa-tail}

\usepackage{hyperref}
\usepackage{url}

\title{The Open Proof Corpus: A Large-Scale Study of LLM-Generated Mathematical Proofs}

\iclrfinalcopy

\author{%
    Jasper Dekoninck\textsuperscript{1,}\thanks{Equal contribution.}\hspace{1.5mm},
    Ivo Petrov\textsuperscript{2,$\ast$},
    Kristian Minchev\textsuperscript{2},
    Mislav Balunović\textsuperscript{1,2},
    Martin Vechev\textsuperscript{1,2} \\
    \textsuperscript{1}ETH Zurich \quad
    \textsuperscript{2}INSAIT, Sofia University "St. Kliment Ohridski" \\
    \texttt{jasper.dekoninck@inf.ethz.ch,ivo.petrov@insait.ai} \\
	\AND
	Dataset Contributors: Miroslav Marinov\textsuperscript{3},
	Maria Drencheva\textsuperscript{2},
	Lyuba Konova\textsuperscript{4}\\
	\textbf{Milen Shumanov\textsuperscript{2}
	Kaloyan Tsvetkov\textsuperscript{2},
	Nikolay Drenchev\textsuperscript{2},
	Lazar Todorov\textsuperscript{2}}\\
	\textbf{Kalina Nikolova\textsuperscript{2,5},
	Nikolay Georgiev\textsuperscript{2},
	Vanesa Kalinkova\textsuperscript{2},
	Margulan Ismoldayev\textsuperscript{2,5}} \\
	\textsuperscript{3}Institute of Mathematics and Informatics, Bulgarian Academy of Sciences \\
	\textsuperscript{4}Sofia University "St. Kliment Ohridski" \quad
	\textsuperscript{5}Massachusetts Institute of Technology
}

\begin{document}

\maketitle
\vspace{-0.6cm}
\begin{centering}
	\includegraphics[height=1em]{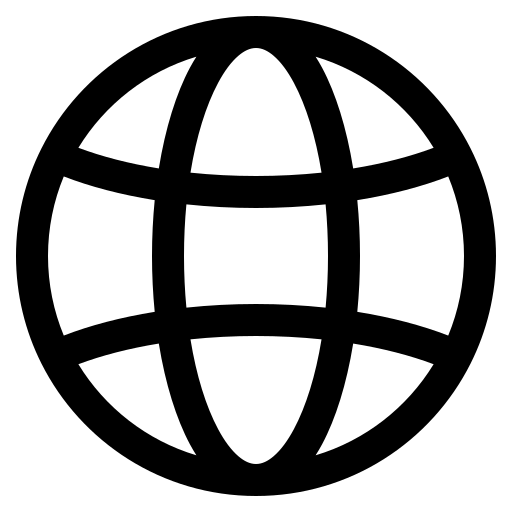} \url{https://proofcorpus.ai/}
	
	\hspace{2.5cm}\includegraphics[height=1em]{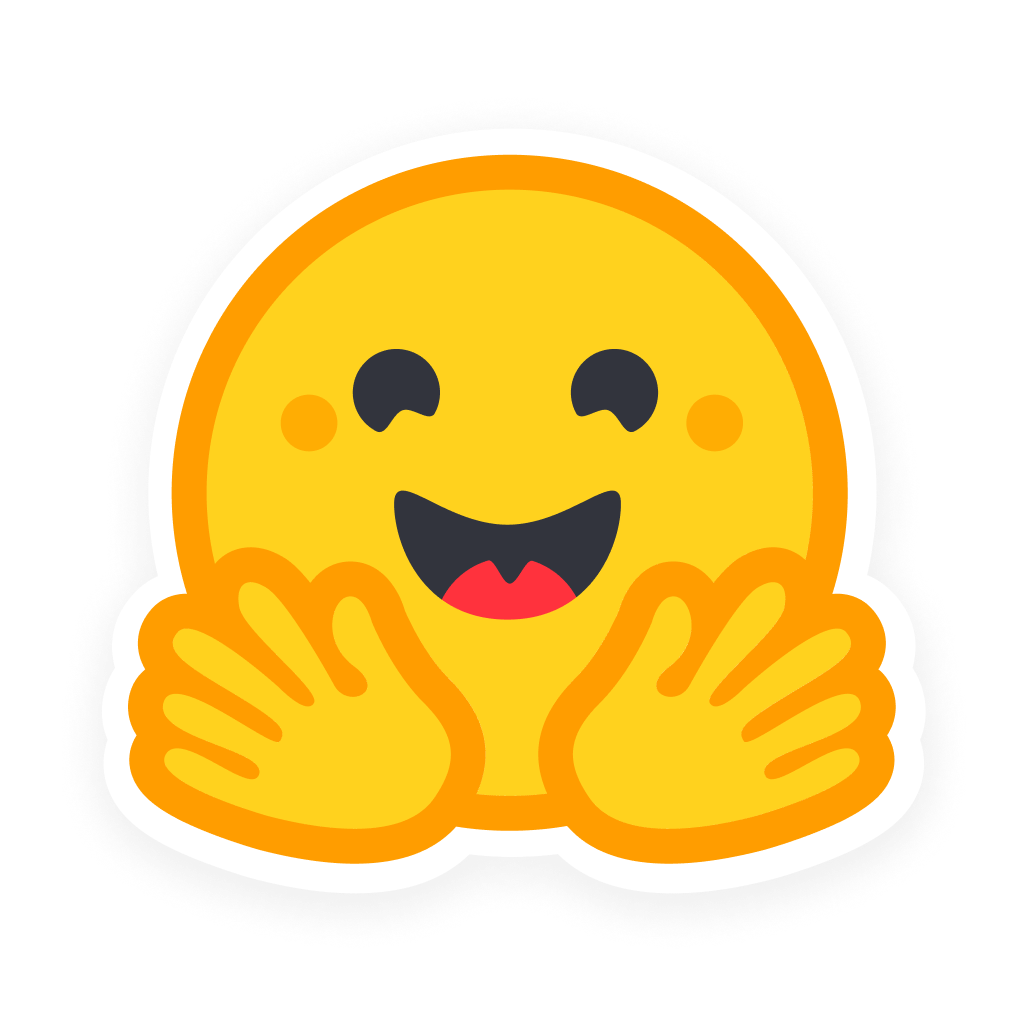} \url{https://huggingface.co/datasets/INSAIT-Institute/OPC}
  \end{centering}
\input{paper_files/abstract}

\input{paper_files/introduction}

\input{paper_files/related.tex}

\input{paper_files/methodology.tex}

\input{paper_files/openproof.tex}

\input{paper_files/results}

\input{paper_files/limitations.tex}

\input{paper_files/conclusion}

\input{paper_files/acknowledgements}
\message{^^JLASTBODYPAGE \thepage^^J}

\bibliography{references}
\bibliographystyle{plainnat}

\message{^^JLASTREFERENCESPAGE \thepage^^J}

\ifbool{includeappendix}{%
	\clearpage
	\appendix
	\onecolumn
	\input{paper_files/appendix}

}{}

\message{^^JLASTPAGE \thepage^^J}

\end{document}

%% file: headers/math_commands.tex
\usepackage{amsmath,amsfonts,bm}

\def\eqref#1{equation~\ref{#1}}

\def\1{\bm{1}}

\DeclareMathAlphabet{\mathsfit}{\encodingdefault}{\sfdefault}{m}{sl}
\SetMathAlphabet{\mathsfit}{bold}{\encodingdefault}{\sfdefault}{bx}{n}

%% file: headers/header-lqa.tex
\usepackage[utf8]{inputenc} %
\usepackage{lipsum} %

\usepackage[T1]{fontenc}

\usepackage{etoolbox}
\newbool{includeappendix}
\setbool{includeappendix}{true}

\input{headers/lqa/overfull}

\input{headers/lqa/comments}
\input{headers/lqa/colors}

\input{headers/lqa/listing}

%% file: headers/lqa/overfull.tex
\ifdefined\isoverfull
\else
\fi

%% file: headers/lqa/colors.tex
\usepackage{xcolor} %

\definecolor{my-full-blue}{HTML}{1F77B4}

\definecolor{my-full-orange}{HTML}{FF7F0E}

\definecolor{my-full-green}{HTML}{2CA02C}

\definecolor{my-full-red}{HTML}{d62728}

\definecolor{my-full-purple}{HTML}{9467bd}

\colorlet{my-blue}{my-full-blue!30}
\colorlet{my-orange}{my-full-orange!30}
\colorlet{my-green}{my-full-green!30}
\colorlet{my-red}{my-full-red!30}
\colorlet{my-purple}{my-full-purple!30}

%% file: headers/lqa/listing.tex
\usepackage{listings}

\usepackage{textcomp}

\usepackage{xcolor}

\usepackage[scaled=0.8]{beramono}

\definecolor{ckeyword}{HTML}{7F0055}
\definecolor{ccomment}{HTML}{3F7F5F}
\definecolor{cstring}{HTML}{2A0099}

\lstdefinestyle{numbers}{
	numbers=left,
	framexleftmargin=20pt,
	numberstyle=\tiny,
	firstnumber=auto,
	numbersep=1em,
	xleftmargin=2em
}

\lstdefinestyle{layout}{
	frame=none,
	captionpos=b,
}

\lstdefinestyle{comment-style}{
	morecomment=[l]//,
	morecomment=[s]{/*}{*/},
	commentstyle={\color{ccomment}\itshape},
}

\lstdefinestyle{string-style}{
	morestring=[b]",%
	morestring=[b]',%
	stringstyle={\color{cstring}},
	showstringspaces=false,%
}

\lstdefinestyle{keyword-style}{
	keywordstyle={\ttfamily\bfseries},
	morekeywords={
		function,
		constructor,
		int,
		bool,
		return,
		returns,
		uint
	},
	morekeywords = [2]{},
	keywordstyle = [2]{\text},
	sensitive=true,
}

\lstdefinestyle{input-encoding}{
	inputencoding=utf8,
	extendedchars=true,
	literate=
	{ℝ}{$\reals$}1%
	{→}{$\rightarrow$}1%
	{α}{$\alpha$}1%
	{β}{$\beta$}1%
	{λ}{$\lambda$}1%
	{θ}{$\theta$}1%
	{ϕ}{$\phi$}1%
}

\lstdefinestyle{escaping}{
	moredelim={**[is][\color{blue}]{\%}{\%}},
	escapechar=|,
	mathescape=true
}

\lstdefinestyle{default-style}{
	basicstyle=\fontencoding{T1}\ttfamily\footnotesize,
	style=numbers,
	style=layout,
	style=comment-style,
	style=string-style,
	style=keyword-style,
	style=input-encoding,
	style=escaping,
	tabsize=2,
	upquote=true
}

\lstdefinelanguage{BASIC}{
	language=C++,
	style=default-style
}[keywords,comments,strings]%

%% file: headers/header.tex
\usepackage{booktabs}       %
\usepackage{nicefrac}       %
\usepackage[flushleft]{threeparttable}

\usepackage{url}
\usepackage{xcolor}
\usepackage{xspace}
\usepackage{etoolbox}
\usepackage{fontawesome}
\usepackage{enumitem}
\usepackage{amsmath,amssymb}
\usepackage{array}
\usepackage{multirow}
\usepackage{wrapfig}
\usepackage{adjustbox,booktabs}
\usepackage{siunitx}
\usepackage{caption}
\usepackage{fontawesome}
\usepackage{stackengine}
\usepackage{svg}
\usepackage{mathtools}
\usepackage{xspace}
\usepackage{wrapfig}
\usepackage{makecell}
\usepackage{graphicx}
\usepackage{booktabs}
\usepackage{longtable}
\usepackage{hyperref}
\usepackage[labelformat=simple]{subcaption}
\usepackage[most]{tcolorbox}
\usepackage{diagbox}

\usepackage{colortbl}

\input{headers/lqa/tikz}

\usepackage{amsthm}
\usepackage{stmaryrd}

\newcolumntype{x}[2]{S[table-format=#1.#2,table-auto-round]}
\newcolumntype{M}{>{$}l<{$}}

\NewDocumentCommand{\phimodel}{O{}}{%
\textsc{Phi}\ifstrempty{#1}{}{-{#1}}\xspace
}

\newcommand{\qwenthreebig}{\textsc{Qwen3-235B-A22B}\xspace}

\newcommand{\ofour}{\textsc{o4-mini}\xspace}
\newcommand{\othree}{\textsc{o3}\xspace}

\newcommand{\rone}{\textsc{R1}\xspace}

\newcommand{\ronesmall}{\textsc{R1-Qwen3-8B}\xspace}
\newcommand{\opcrone}{\textsc{OPC-R1-8B}\xspace}

\newcommand{\grokfour}{\textsc{Grok-4}\xspace}
\newcommand{\grokthreemini}{\textsc{Grok-3-Mini}\xspace}

\newcommand{\claude}{\textsc{Claude-4-Sonnet}\xspace}
\newcommand{\geminipro}{\textsc{Gemini-2.5-Pro}\xspace}
\newcommand{\geminiflash}{\textsc{Gemini-2.5-Flash}\xspace}
\newcommand{\gptfive}{\textsc{GPT-5}\xspace}

\definecolor{acceptblue}{HTML}{6494EA}
\hypersetup{citecolor=acceptblue}

\definecolor{lightred}{HTML}{ffcbc7}
\definecolor{gemini}{HTML}{4285F4}
\definecolor{claude}{HTML}{f3e9d7}
\definecolor{deepseek}{HTML}{FADA4B}
\definecolor{qwen}{HTML}{FA574B}
\definecolor{oai}{HTML}{10a37f}
\definecolor{LightGreen}{HTML}{CCFFCC}

\lstdefinestyle{mystyle}{
    breaklines=true,
    basicstyle=\scriptsize\ttfamily,
    numbers=none,
    language={},
    framextopmargin=0pt,
    framexbottommargin=0pt,
    breakindent=0pt,
    showspaces = false,
    keywordstyle=\bfseries,
    showstringspaces=false,
    columns=fullflexible,
    morekeywords={Answer}
    moredelim=[**][\bfseries]{!!}
}

\newtcblisting{prompt}[2][]{
    arc=3pt, outer arc=3pt,
    width=\linewidth,
    left=1mm,
    top=0mm,
    bottom=0mm,
    title=#2, 
    colback=lightred!5!white,
    colframe=black,
    fonttitle=\bfseries,
    listing only, 
    listing options={style=mystyle},
    breakable,
    #1
}

\newtcblisting{groksl}[2][]{
    arc=3pt, outer arc=3pt,
    width=\linewidth,
    left=1mm,
    top=0mm,
    bottom=0mm,
    title=#2, 
    colback=lightred!5!white,
    colframe=black,
    fonttitle=\bfseries,
    listing only, 
    listing options={style=mystyle},
    breakable,
    #1
}

\newtcblisting{problem}[2][]{
    arc=3pt, outer arc=3pt,
    width=\linewidth,
    left=1mm,
    top=0mm,
    bottom=0mm,
    title=#2, 
    colback=acceptblue!5!white,
    colframe=acceptblue!50!black,
    fonttitle=\bfseries,
    listing only, 
    listing options={style=mystyle},
    breakable,
    #1
}

\newtcblisting{gpt}[2][]{
    arc=3pt, outer arc=3pt,
    width=\linewidth,
    left=1mm,
    top=0mm,
    bottom=0mm,
    title=#2, 
    colback=oai!5!white,
    colframe=oai!50!black,
    fonttitle=\bfseries,
    listing only, 
    listing options={style=mystyle},
    breakable,
    #1
}

\newtcblisting{claudels}[2][]{
    arc=3pt, outer arc=3pt,
    width=\linewidth,
    left=1mm,
    top=0mm,
    bottom=0mm,
    title=#2, 
    colback=claude!5!white,
    colframe=claude!85!black,
    fonttitle=\bfseries,
    listing only, 
    listing options={style=mystyle},
    breakable,
    #1
}

\newtcblisting{deepseekls}[2][]{
    arc=3pt, outer arc=3pt,
    width=\linewidth,
    left=1mm,
    top=0mm,
    bottom=0mm,
    title=#2, 
    colback=deepseek!5!white,
    colframe=deepseek!85!black,
    fonttitle=\bfseries,
    listing only, 
    listing options={style=mystyle},
    breakable,
    #1
}

\newtcblisting{geminils}[2][]{
    arc=3pt, outer arc=3pt,
    width=\linewidth,
    left=1mm,
    top=0mm,
    bottom=0mm,
    title=#2, 
    colback=gemini!5!white,
    colframe=gemini!50!black,
    fonttitle=\bfseries,
    listing only, 
    listing options={style=mystyle},
    breakable,
    #1
}

\newtcblisting{geminilsescaped}[2][]{
    arc=3pt, outer arc=3pt,
    width=\linewidth,
    left=1mm,
    top=0mm,
    bottom=0mm,
    title=#2, 
    colback=gemini!5!white,
    colframe=gemini!50!black,
    fonttitle=\bfseries,
    listing only, 
    listing options={style=mystyle, escapeinside=||},
    breakable,
    #1
}

\newtcblisting{qwenls}[2][]{
    arc=3pt, outer arc=3pt,
    width=\linewidth,
    left=1mm,
    top=0mm,
    bottom=0mm,
    title=#2, 
    colback=qwen!5!white,
    colframe=qwen!50!black,
    fonttitle=\bfseries,
    listing only, 
    listing options={style=mystyle},
    breakable,
    #1
}

\newcommand{\geminiproheader}{
    \begin{tikzpicture}
      \node[anchor=north] {\pgftext{\includegraphics[width=0.34cm]{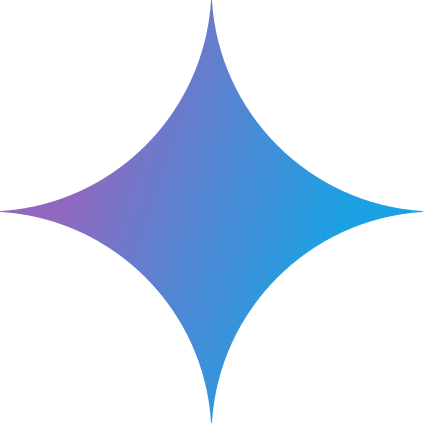}}};
    \end{tikzpicture}
    ~\geminipro{}
}

\newcommand{\deepseekheader}{
    \begin{tikzpicture}
      \node[anchor=north] {\pgftext{\includegraphics[width=0.34cm]{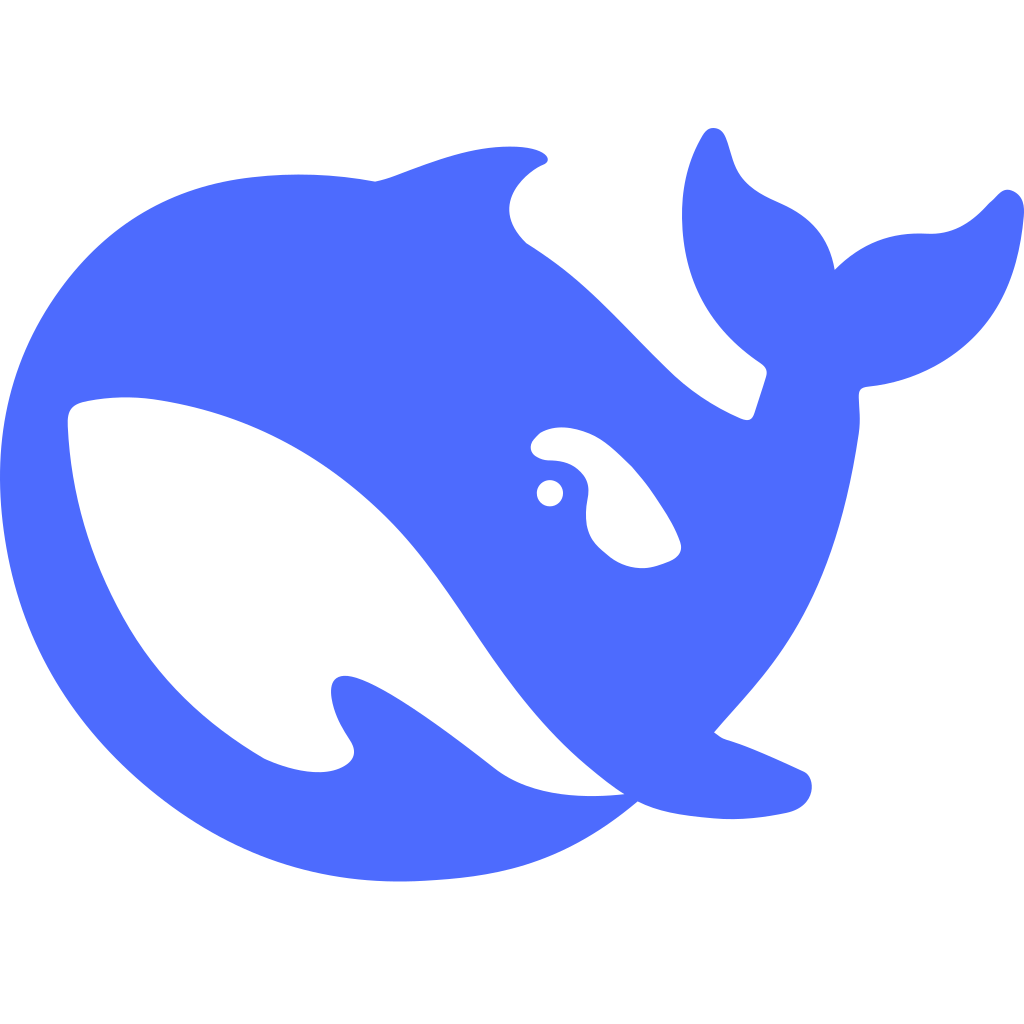}}};
    \end{tikzpicture}
    ~\rone{} 
}

\newcommand{\grokheader}{
    \begin{tikzpicture}
      \node[anchor=north] {\pgftext{\includegraphics[width=0.34cm]{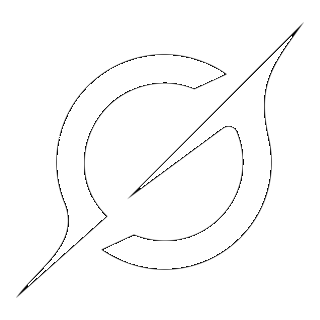}}};
    \end{tikzpicture}
    ~\textsc{Grok 3 Mini}
}

%% file: headers/lqa/tikz.tex
\usepackage{tikz}

\usetikzlibrary{calc,decorations,decorations.pathmorphing}
\usetikzlibrary{positioning,fit,arrows}
\usetikzlibrary{decorations.markings}
\usetikzlibrary{shapes,shapes.geometric}
\usetikzlibrary{shadows,patterns,snakes}
\usetikzlibrary{backgrounds,decorations.pathreplacing,calligraphy,automata}
\usetikzlibrary{intersections}
\usetikzlibrary{angles,quotes}
\usetikzlibrary{plotmarks}
\usetikzlibrary{patterns}
\usetikzlibrary{arrows.meta}
\usetikzlibrary{shapes.misc}
\usetikzlibrary{chains}
\usetikzlibrary{shapes.callouts}

%% file: headers/header-lqa-tail.tex
\input{headers/lqa/references}

%% file: headers/lqa/references.tex
\usepackage[capitalize]{cleveref}

\crefformat{section}{\S#2#1#3}

\crefrangeformat{section}{\S#3#1#4\crefrangeconjunction\S#5#2#6}

\crefmultiformat{section}{\S#2#1#3}{\crefpairconjunction\S#2#1#3}{\crefmiddleconjunction\S#2#1#3}{\creflastconjunction\S#2#1#3}

\newcommand{\crefrangeconjunction}{--}

\crefname{listing}{Lst.}{listings}
\crefname{line}{Lin.}{Lin.}
\crefname{appendix}{App.}{App.}

\newcommand{\app}[1]{%
	\ifbool{includeappendix}{\cref{#1}}{the appendix}%
}
\newcommand{\App}[1]{%
	\ifbool{includeappendix}{\cref{#1}}{The appendix}%
}

%% file: paper_files/abstract.tex
\begin{abstract}
    In recent months, large language models (LLMs) have made significant progress in mathematical proof generation, but further advancement is hindered by the lack of a large-scale, high-quality dataset of human-evaluated proofs. While expensive to create, such a dataset is essential for driving improvements in training and addressing key open questions in the field of automated proof generation. Specifically, it remains unknown (1) how large the gap is between natural language and formal proof generation, (2) how final-answer accuracy relates to full proof correctness, and (3) how best-of-n selection strategies can affect proof quality.
    In this work, we present \emph{the Open Proof Corpus} (OPC), a dataset comprising over 5,000 human-evaluated proofs produced by state-of-the-art LLMs. The OPC was specifically designed for broad applicability and downstream usage in proof generation research and is the first large dataset of LLM-generated solutions to problems from prestigious mathematics competitions such as the USAMO and IMO. Using the OPC, we address the open questions outlined above and provide new insights into LLMs' strengths and limitations in mathematical reasoning. Finally, to showcase the utility of the OPC, we finetune an 8B-parameter model on the dataset, obtaining a model that matches \geminipro{}, and performs close to the best model, \gptfive{}, on evaluating proof correctness.
\end{abstract}

%% file: paper_files/introduction.tex
\section{Introduction}\label{sec:intro}

Large language models (LLMs) have recently achieved impressive progress in mathematical reasoning, obtaining top-competitor performance on various final-answer benchmarks such as AIME and HMMT \citep{matharena}. However, growing evidence suggests that these benchmarks fail to capture the full range of mathematical capabilities, as they do not require models to produce proofs or detailed intermediate steps \citep{brainvbytes,rightisnotenough}. Such step-by-step reasoning is critical for applications in theorem proving, mathematical research, and education.
\vspace{1mm}
\paragraph{Proof benchmarking} To address this, several evaluation efforts have taken place, revealing that LLMs significantly underperform on proof generation compared to existing final-answer benchmarks \citep{prooforbluff,brainvbytes}. Despite their value, these benchmarks are severely limited in their use for broader analysis and training. Specifically, they are small \citep{prooforbluff,rightisnotenough}, use outdated models \citep{frieder2024}, contain few correct proofs \citep{brainvbytes}, and are not open-sourced \citep{brainvbytes,rightisnotenough}.
\vspace{1mm}
\paragraph{Open questions} Furthermore, key questions about proof generation capabilities remain unanswered. First, while it is widely claimed that there is a gap between final-answer performance and proof generation capabilities, this claim has not yet been supported by evaluating LLM-generated proofs on existing final-answer benchmarks. Second, despite recent advances in formal proof generation with Lean \citep{seedprover,goedelproverv2}, the performance gap between natural language and formal proof generation remains unclear. Third, the potential of best-of-n selection strategies to improve proof quality has not been explored.

\begin{figure*}[t]
    \centering
    \begin{subfigure}[t]{0.6\textwidth}
      \centering
      \resizebox{\linewidth}{!}{
        \input{figures/overview_sample.tex}
      }
      \caption{Sample from the OPC}
      \label{fig:overview:left}
    \end{subfigure}\hfill%
    \begin{subfigure}[t]{0.38\textwidth}
      \centering
      \includegraphics[width=\linewidth]{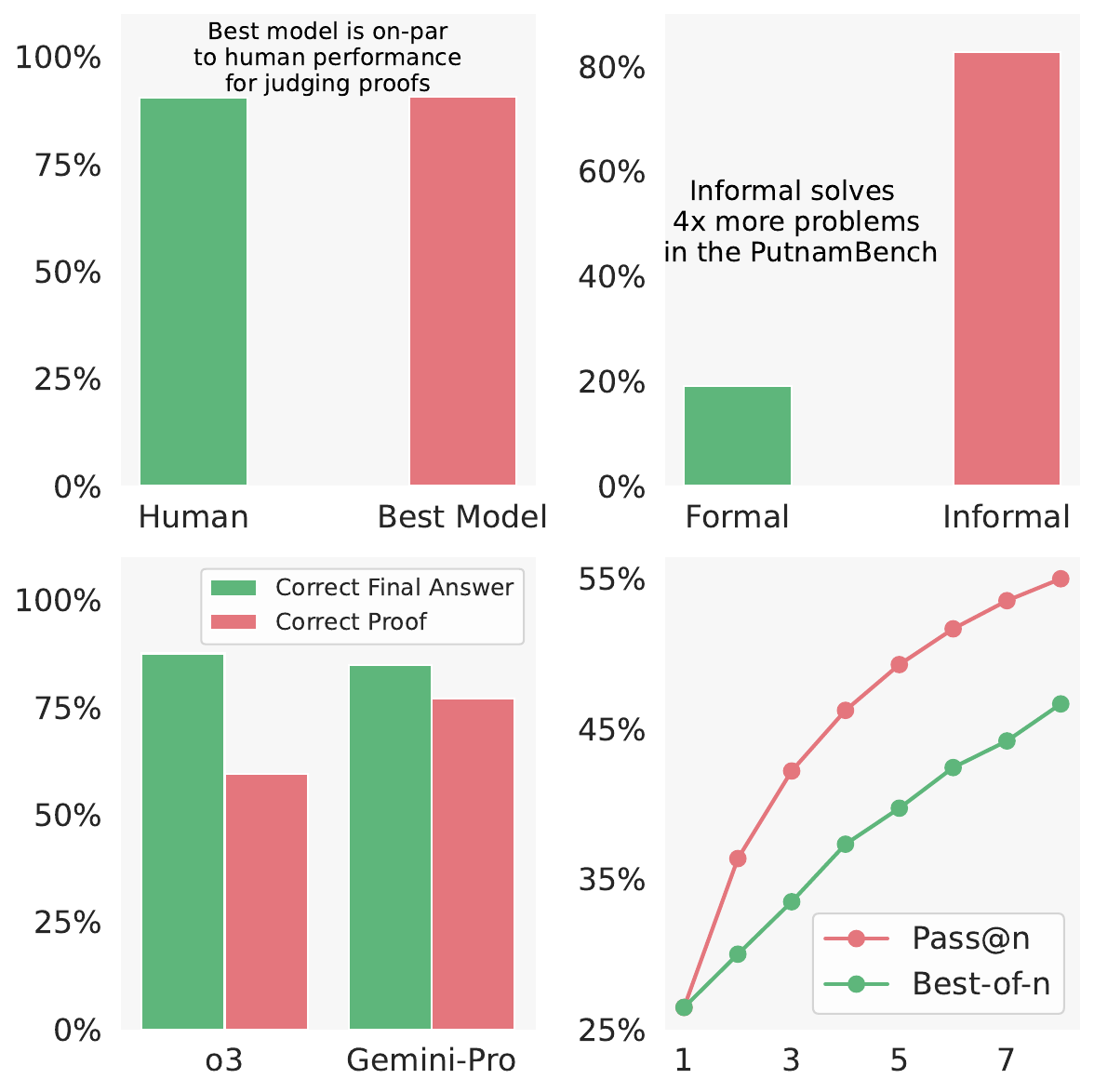}
      \caption{Conclusions from the OPC}
      \label{fig:overview:right}
    \end{subfigure}
    \vspace{-2mm}
    \caption{Overview of the OPC and its conclusions. On the left, we show a typical sample, including a question from a high-quality mathematical competition, an LLM-generated proof, and a human judgment of proof correctness. On the right, we summarize the main conclusions from the OPC.}
    \vspace{-3mm}
    \label{fig:overview}
  \end{figure*}

  \vspace{-1mm}
\paragraph{Our work: the Open Proof Corpus} To address these challenges, we introduce the \textit{Open Proof Corpus (OPC)}: a large-scale, human-validated dataset comprising over $5{,}000$ LLM-generated proofs across more than $1{,}000$ problems. As shown in \cref{fig:overview:left}, each OPC sample includes (1) a problem from a high-quality mathematical competition such as the International Mathematical Olympiad, (2) a proof generated by a state-of-the-art LLM, and (3) a binary human evaluation of the proof's correctness with feedback. The OPC was specifically designed for downstream usage in proof generation research, enabling both training and evaluation of LLMs on proof generation tasks. To address the open questions outlined earlier, the OPC includes problems from specific sources: problems from the PutnamBench \citep{putnambench} enable comparisons between formal and informal reasoning, while problems from MathArena \citep{matharena} support evaluation of proof correctness for problems with final answers.

\paragraph{Key findings} Despite the difficulty of problems in the OPC, state-of-the-art LLMs demonstrate surprisingly strong performance. For instance, \ofour{} correctly solves almost $20\%$ of the problems in the IMO Shortlist, and the OPC generally consists of $43\%$ correct proofs. Furthermore, as shown in the top left of \cref{fig:overview:right}, LLMs exhibit strong capabilities in evaluating proofs: \gptfive{} achieves $90.8\%$ accuracy in judging proof correctness, which is on-par with human performance on this task. To showcase the utility of the OPC, we fine-tune \ronesmall{} \citep{r1} using GRPO \citep{grpo} on the OPC, resulting in an open model that achieves $88.1\%$ judgment accuracy, close to top models like \gptfive{} and outperforming the majority of frontier models. 

\vspace{-1mm}
\paragraph{Answering open questions} The OPC allows us to empirically resolve the open questions outlined above, with all conclusions shown in \cref{fig:overview:right}.  First, as shown on the top right, we find that natural language proof generation significantly outperforms formal proof generation: on PutnamBench, \geminipro{} solves 4 times more problems than the best formal model, \textsc{Goedel-Prover-V2} \citep{goedelproverv2}. Second, as shown on the bottom left, we observe a substantial gap between final-answer accuracy and proof correctness. While \geminipro{} loses only $8\%$ of its final-answer accuracy when proof correctness is required, \othree{} suffers a drop of almost $30\%$. Third, as shown on the bottom right, we find that best-of-n strategies yield large gains in performance. Interestingly, while standard best-of-n selection methods moderately improve accuracy from $26\%$ to $36\%$, a ranking approach inspired by \citep{knockout} achieves the highest performance of $47\%$.

\vspace{-1mm}
\paragraph{Main Contributions} Our key contributions are:
\vspace{-2mm}
\begin{itemize}\setlength\itemsep{0.005em}
    \item A rigorous pipeline for generating and evaluating natural language proofs (\cref{sec:methodology}).
    \item The OPC, a human-validated dataset comprising over 5,000 LLM-generated proofs (\cref{sec:openproof}).
    \item Resolution of open questions in natural language proof generation using the OPC (\cref{sec:results}).
    \item An open-source, 8B-parameter model fine-tuned on the OPC that achieves $88.1\%$ judgment accuracy, close to the best model on this task (\cref{sec:results}).
\end{itemize}

%% file: figures/overview_sample.tex
\begin{tikzpicture}[
    node distance=1cm and 2cm,
    box/.style = {draw=blue!80, fill=blue!10, rectangle, minimum width=2.5cm, minimum height=0.5  cm, align=center},
    >=stealth
  ]
  \definecolor[grey]{grey}{RGB}{190, 190, 190}
  \definecolor[lightgrey]{lightgrey}{RGB}{240, 240, 240}
  \definecolor{darkgreen}{RGB}{100,170,110} %
  \definecolor{lightgreen}{HTML}{5eb67b} %
  \definecolor{darkblue}{RGB}{100, 170, 250} %
  \definecolor{lightblue}{RGB}{180, 250, 250} %
  \definecolor{darkred}{RGB}{250, 100, 100} %
  \definecolor{redcolor}{HTML}{e4767d} %
  \definecolor{refmodelcolor}{RGB}{222, 143, 5} 

  \tikzstyle{box} = [rectangle, rounded corners, fill=lightgray!50, text centered, text width=6cm, font=\large, minimum height=2.5cm, font=\LARGE]

  \tikzstyle{arrow} = [->, thick, >=stealth, line width=1mm, color=grey]

  \tikzstyle{smallarrow} = [->, >=stealth, line width=0.4mm, color=grey]

  \tikzstyle{dashedline} = [draw=lightgray, thick, dash pattern=on 0pt off 3pt on 5pt off 3pt, line width=0.5mm]

  \tikzstyle{mainNode} = [
      rectangle, 
      text=white, 
      fill=lightgrey, 
      minimum height=4cm, 
      minimum width=6.5cm, 
      align=center, 
        draw=black,
      rounded corners
  ]
  \tikzstyle{subTitleNode} = [
    rectangle,
    fill=darkgreen,
    align     = left, 
    text width = 3cm,   %
    minimum height=0.3cm, %
    minimum width=2.9cm,  %
    inner sep=0,
    font=\fontsize{6pt}{6.5pt}\selectfont,
    anchor=west,
  ]

  \tikzstyle{subNode} = [
    rectangle,
    fill=lightgreen,
    align     = left, 
    minimum height=0.45cm, %
    minimum width=3cm,  %
    inner sep=0,
    font=\fontsize{5pt}{5.5pt}\selectfont,
    text width = 2.9cm,
    align = left,  
    anchor=west,
  ]

  \tikzstyle{parsingSubNode} = [
    rectangle,
    fill=lightgreen,
    draw, 
    minimum height=0.35cm, %
    minimum width=0.95cm,  %
    inner sep=0,
    font=\fontsize{7pt}{7.5pt}\selectfont,
    rounded corners,
  ]

    \node[mainNode] (sample3) {};
    \node[mainNode, anchor=north west, xshift=-0.1cm, yshift=0.1cm] (sample2) at (sample3.north west) {};
    \node[mainNode, anchor=north west, xshift=-0.1cm, yshift=0.1cm] (sample) at (sample2.north west) {};
    \node[anchor=north west, font=\tiny, text=gray] at (sample.north west) {\textit{Sample}};

    \node[rectangle, draw, fill=grey, minimum width=5.8cm, text width=5.8cm, font=\tiny, align=left,rounded corners, anchor=north, yshift=-0.4cm] (metadata) at (sample.north) {\textbf{Problem}: EGMO 2012, Problem 3\\
\textbf{Level}: High School\\
\textbf{Split}: Generic
    };
    
    \node[rectangle, draw, fill=darkblue, minimum width=5cm, text width=5cm, font=\tiny, align=center,rounded corners, anchor=north, yshift=-0.15cm, xshift=0.4cm] (question) at (metadata.south) {\textbf{Question}: Solve over \(\mathbb{R}\) the functional equation $f(y f(x+y)+f(x))=4 x+2 y f(x+y)$};

    \node[anchor=east, xshift=-0cm] (human1) at (question.west) {
        \includegraphics[width=0.7cm]{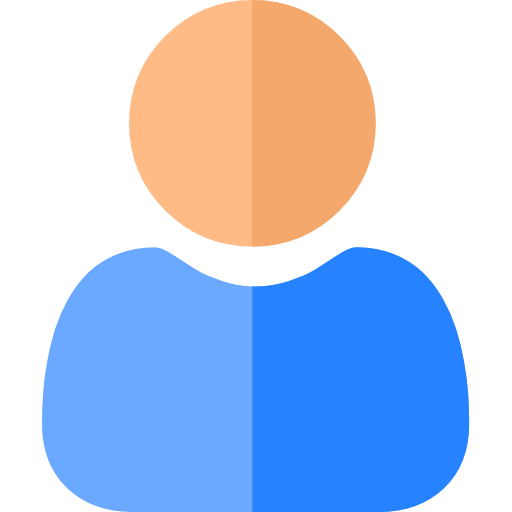}
    };

    \node[rectangle, draw, fill=lightgreen, minimum width=5cm, text width=5cm, font=\tiny, align=center,rounded corners, anchor=north, yshift=-0.15cm, xshift=-0.8cm] (answer) at (question.south) {\textbf{Proof}: To solve the functional equation \dots we conclude that this is the **only** solution.};

    \node[anchor=west, xshift=-0cm] (bot) at (answer.east) {
        \includegraphics[width=0.7cm]{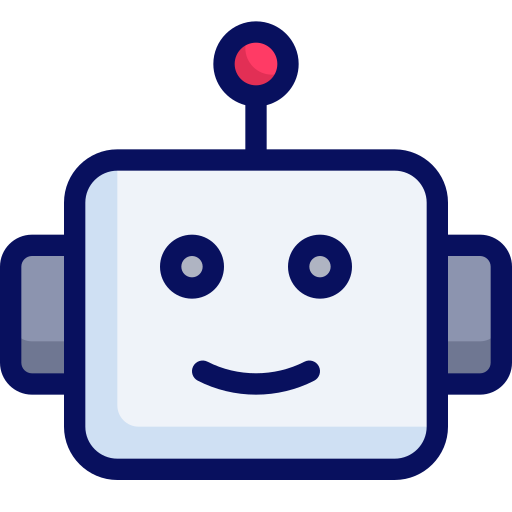}
    };

    \node[rectangle, draw, fill=redcolor, minimum width=5cm, text width=5cm, font=\tiny, align=center,rounded corners, anchor=north, yshift=-0.15cm, xshift=0.8cm] (judgment) at (answer.south) {\textbf{Grade (Incorrect)}: The model substitutes $x$ to be 1. All further ideas are based on the wrong assumption.};

    \node[anchor=east, xshift=-0cm] (human2) at (judgment.west) {
        \includegraphics[width=0.7cm]{figures/man.png}
    };

\end{tikzpicture}

%% file: paper_files/related.tex
\vspace{-2mm}
\section{Related Work}
\label{sec:related}
We briefly discuss the relevant literature on mathematical reasoning benchmarks and datasets. 

\paragraph{Final-answer benchmarks} Final-answer benchmarks evaluate models by comparing a final answer from the model's output with a ground-truth answer, typically using rule-based parsers. With the rise of reasoning LLMs \citep{o1}, older benchmarks have become saturated \citep{gsm8k,math500}, and more recent ones are nearing saturation \citep{omnimath,olympiadbench,matharena}. Only private benchmarks such as FrontierMath \citep{frontiermath} remain sufficiently challenging for the latest models. However, this benchmark does not require the generation of full proofs or detailed reasoning steps. Moreover, its private nature hinders reproducibility and broader community engagement, both key factors in driving progress.

\paragraph{Formal proof generation} Another growing line of work involves training LLMs to generate formal proofs in languages such as Lean \citep{lean} or Isabelle \citep{isabelle}, which can then be automatically verified by these systems \citep{minif2f,putnambench,formalmath}. While this paradigm enables rigorous correctness checking, it typically requires models to be specifically finetuned for formal proof generation \citep{deepseekproverv2,kimina,goedelprover}. In contrast, state-of-the-art general-purpose models like \gptfive{} and \geminipro{} struggle with formal proof generation. As a result, formal reasoning currently does not make full use of the natural language capabilities of these general-purpose models, and, as we show in \cref{sec:results}, there remains a significant performance gap between formal and natural language proof generation, with the latter being substantially more effective.

\paragraph{Proof-generation evaluation efforts} Going beyond final-answer accuracy has gained recent attention, with several works investigating the reasoning traces of recent LLMs to identify patterns and potential for improvement \citep{illusionthinking,beyondaccuracysurvey,llmjudgebeyondaccuracy}. However, only a few studies have focused directly on evaluating full proofs. \citet{prooforbluff} evaluated LLMs on the six problems from the USAMO 2025, uncovering significant flaws in the generated proofs. Similarly, \citet{brainvbytes} evaluated LLM performance on a large set of IMO Shortlist problems, finding that no model surpassed $5\%$ accuracy. In contrast, \citet{frieder2024} reported that \textsc{GPT-3.5} and \textsc{GPT-4} performed well on simpler tasks, generating correct proofs in a significant fraction of cases. Still, all these studies are limited by either the use of outdated models or the small scale of their evaluations.

Two recent works have expanded this line of research. \citet{ineqmath} focus specifically on inequality proofs and primarily emphasize the development of an LLM as a judge framework to mitigate the high cost of human evaluation. \citet{rightisnotenough} highlight a notable gap between final-answer accuracy and the ability to generate correct proofs, a finding we confirm in \cref{sec:results}. However, their analysis stops at this observation and does not evaluate performance on an established final-answer benchmark. Further, both studies are limited in scale compared to the OPC and have not open-sourced their human-annotated proofs.

\paragraph{Automated proof grading with LLMs as judges}
The growing capabilities of LLMs have led to the ``LLM-as-a-judge'' paradigm, which enables cost-effective evaluation of complex model outputs \citep{llmasajudge,llmasjudgesurvey,chevalier2024science}. Recent work has extended this approach to mathematical proof evaluation \citep{ineqmath,zhao2024autograding,zhao2025fewshotgraders}. These studies report promising results, showing that LLMs can grade specific types of problems with reasonable reliability \citep{ineqmath,zhao2024autograding} and can provide useful feedback in educational settings \citep{zhao2024autograding,zhao2025fewshotgraders}. This work builds on these insights, building a dataset of human-evaluated proofs to further improve LLM judging capabilities, as detailed in \cref{sec:results}.

\paragraph{Mathematical training datasets} Several large-scale datasets have been developed to train LLMs on mathematical reasoning. One of the earliest, \citet{numina_math_datasets}, compiled a large dataset of internet-sourced problems, including both final-answer questions and natural language proofs. However, it lacks LLM-generated proofs, human judgments, and examples of incorrect proofs. Other datasets have focused exclusively on final-answer tasks \citep{bigmath,deepmath,aimo2}, offering limited support for training or evaluating proof generation. Finally, \citet{deeptheorem} introduced a dataset of both valid and invalid problem statements, each accompanied by LLM-generated proofs. While this ensures that incorrect proofs exist for invalid statements, the dataset does not include human evaluation of proofs or other information on proof validity.

%% file: paper_files/methodology.tex
\section{Methodology}\label{sec:methodology}
Accurately evaluating LLM-generated proofs is a challenging task. Models often make difficult-to-detect errors, and they rarely acknowledge when they cannot solve a problem (see \cref{sec:results}). In this section, we outline the methodology used to create the OPC, with particular emphasis on the complexities of evaluating LLM-generated proofs and our efforts to maximize dataset size. Since human judges cannot reasonably spend hours studying each problem, we developed a pipeline to support efficient grading. This pipeline consisted of three main components: problem and judge preparation (\cref{sec:judgeproblem}), the grading procedure (\cref{sec:grading}), and monitoring and validation (\cref{sec:monitoring}). The dataset was constructed over a period of four weeks, involving $13$ expert judges and generating over $5{,}000$ human-validated proofs.

\subsection{Problem and Judge Preparation} \label{sec:judgeproblem}

\paragraph{Judge selection} Judges were selected from among former IMO participants or individuals who reached the final stages of IMO selection in their countries. We contacted each judge personally to ensure they were qualified and motivated. A total of $13$ judges were involved, each responsible for grading a varying number of problems. Three of the most active judges had prior experience with evaluating LLM-generated proofs. One judge served as the coordinator, maintaining regular communication, tracking progress, and ensuring consistency and motivation across the group. 

\paragraph{Problem selection} Problems were drawn from top-tier national and international mathematics competitions, such as the USAMO and IMO, to capture a balanced mix of correct and incorrect proofs. In particular, we included competitions based on two informal criteria: (1) the competition is well-known and produces high-quality problems, and (2) the difficulty level of the competition aligns with our target of roughly 50\% model accuracy. All problems were sourced from official materials. Non-English problems were translated using \textsc{GPT-4.1} and manually verified for accuracy by the coordinator. When available, official ground-truth solutions were also extracted and provided to the judges as references.

Throughout the annotation process, model performance was actively monitored to ensure that the selected problems remained appropriately challenging. For instance, more problems from international competitions were added when initial results indicated that models were performing very well ($\approx 65\%$) on national-level problems. Each day, problem prioritization was adjusted based on ongoing performance metrics, judge availability, and progress towards the specific conclusions we aimed to draw from the dataset.

\vspace{-1mm}
\paragraph{Proof generation} Proofs were generated using a set of state-of-the-art LLMs known for their strength in mathematical reasoning. Specifically, we used \ofour{} and \othree{} from OpenAI \citep{o4}, \geminipro{} from Google \citep{deepmind2025geminipro}, \grokthreemini{} from xAI \citep{xai2025grok3}, \qwenthreebig{} from Qwen \citep{qwen3blog2025}, and the latest version of \rone{} from DeepSeek \citep{r1}. \rone{} was released mid-way through dataset construction and replaced \grokthreemini{} thereafter. We designed the model prompt to clearly instruct models to generate full solutions, refining it through small-scale pilot tests. The final prompt is shown in \cref{app:proof_generation_prompt}. All models were run with default parameters and a maximum generation length of $64{,}000$ tokens. For the MathArena subset \citep{matharena}, we only retained solutions with a correct final answer, retrying generation if necessary. For PutnamBench \citep{putnambench}, we appended the informal final answer (if present) to the problem statement to mirror the setup for formal models, allowing direct comparison between natural and formal proof outputs.

\vspace{-1mm}
\subsection{Grading Procedure} \label{sec:grading}
\vspace{-1mm}

\paragraph{User interface} We built a custom web interface to facilitate efficient grading. A sample instance is available in our supplementary material, with screenshots included in \cref{app:grading_interface}. The interface displays the problem, reference solution (if available), anonymized model-generated proof, and grading form. Judges could mark the problem or solution as malformed (to filter such cases from the dataset), grade the proof, and annotate sentences from the model-generated proof with comments. Continuous feedback from judges helped us refine the interface over time.

\vspace{-0.5mm}
\paragraph{Judge instructions} Judges were asked to label proofs as either correct or incorrect and provide written justification. Precise grading guidelines were critical to avoid inconsistencies in edge cases, such as minor omissions or shortcuts. To prevent overly strict grading, we clearly defined what level of omissions and frequency of mistakes were acceptable in a correct proof. The full guidelines are available in our supplementary material, with a summary in \cref{app:grading_interface}. These instructions were developed collaboratively with the judges and finalized after a pilot phase (see \cref{sec:monitoring}).

\paragraph{Abstention and uncertainty} Judges were allowed to abstain from grading a proof if they lacked the necessary expertise or found the solution too complex or convoluted. They could also mark judgments as uncertain in borderline cases, which proved especially useful for near-correct proofs containing subtle errors. Less than $3\%$ of proofs in the OPC are flagged as uncertain.

\vspace{-0.5mm}
\paragraph{LLM issue summaries} After several hundred graded proofs, we introduced a new feature to support grading: an automatically generated summary of the proof using \ofour{}. These summaries flagged potential issues, such as logical gaps or missing steps, based on a specially designed prompt (see \cref{app:issues_interface_prompt}). Importantly, the model was instructed not to give a final verdict. Judges reported that this significantly improved their efficiency and accuracy in detecting errors. To ensure the inclusion of these summaries did not bias our judges, we evaluated the agreement rate between \ofour{} as a judge and human graders before and after their introduction. There was no significant difference in agreement, indicating that no bias was introduced into the grading process. However, in experiments involving best-of-n selection, where the LLM judge acts as a selection mechanism, we omitted these summaries to avoid any form of compounding bias in the evaluation.

\vspace{-0.5mm}
\paragraph{Problem distribution} Problems were assigned to judges based on background and availability. Judges not qualified to evaluate undergraduate-level problems were excluded from grading them. To ensure a balanced workload, we monitored grading progress and dynamically reassigned problems as needed.

\vspace{-1mm}
\subsection{Monitoring and Validation} \label{sec:monitoring}
\vspace{-1mm}

To ensure grading quality and judge consistency, we implemented a set of validation procedures.

\vspace{-1mm}
\paragraph{Coordinator} One experienced judge was assigned as coordinator, responsible for tracking grading progress, resolving issues, and ensuring motivation. As a core author, the coordinator had detailed knowledge of the dataset's goals and methodology and was available to answer judges' questions.

\paragraph{Developer} In addition to the coordinator, one of the core authors served as developer, providing technical support for the grading interface and addressing any software-related issues that arose during the grading process. Judges were encouraged to report bugs or suggest improvements to the grading interface, which the developer would address immediately to ensure a smooth grading experience. The developer was also responsible for implementing new features, such as the LLM-generated issue summaries, and to monitor model performance metrics to inform problem selection and judge assignments.

\vspace{-1mm}
\paragraph{Pilot phase} Before full-scale grading, we conducted a test run with a limited number of problems. In particular, four experienced judges graded approximately 300 proofs in this initial stage. During this phase, a more significant portion of proofs (around $35\%$) were double-graded to evaluate inter-judge consistency. Judges were also very actively monitored, and encouraged to actively ask questions about any small ambiguity they encountered. After this stage, judge feedback was collected to improve the grading interface and instructions. In particular, the instructions were refined to address remaining ambiguities and edge cases. 

\vspace{-1mm}
\paragraph{Monitoring discrepancies} In total, approximately $10\%$ of the proofs were double-graded to evaluate inter-judge consistency. Throughout the grading pipeline, disagreements were reviewed by the coordinator to determine whether they arose from misunderstandings, ambiguous instructions, or errors overlooked by a judge. If possible, instructions were further improved to prevent similar issues. However, most inconsistencies came from overlooked errors in the proofs and could therefore not be resolved by clarifying the instructions. If the coordinator identified a significant number of discrepancies for a specific judge, they would discuss the issue with the judge to clarify instructions.

%% file: paper_files/openproof.tex
\vspace{-1mm}
\section{Open Proof Corpus}\label{sec:openproof}

We now introduce the OPC and provide an overview of key dataset statistics. At a high level, the OPC is built to support training and to facilitate a rigorous analysis of proof generation capabilities.

\vspace{-1mm}
\begin{wrapfigure}[14]{r}{0.5\textwidth} %
    \centering
    \vspace{-5mm}
    \includegraphics[width=0.48\textwidth]{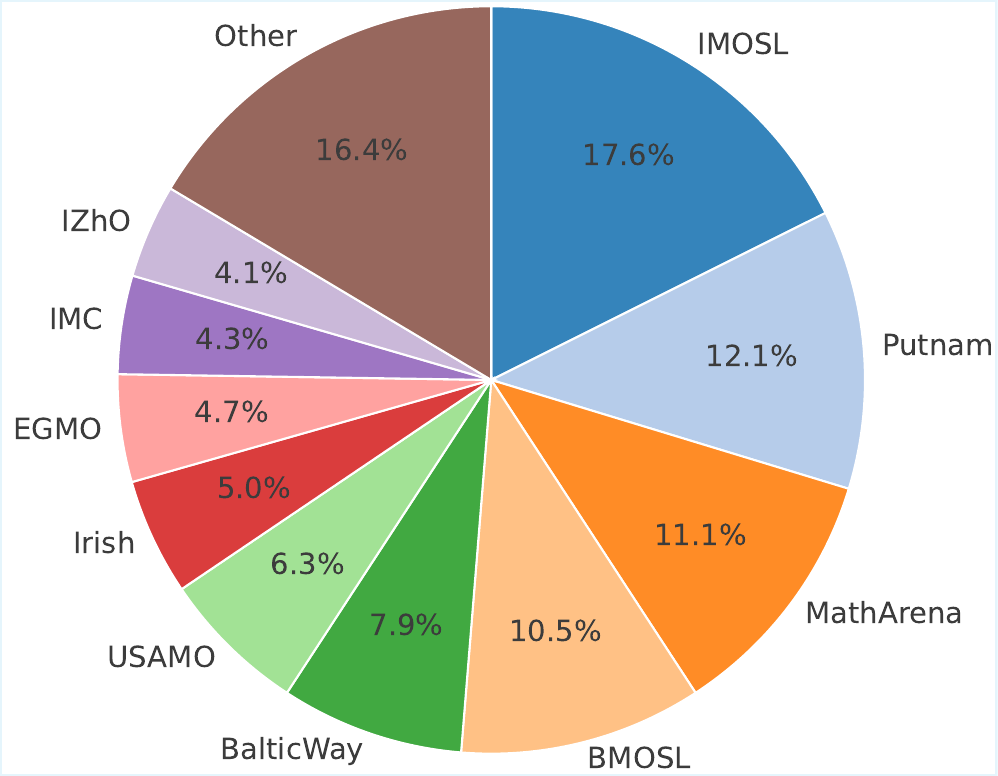}
    \vspace{-3mm}
    \caption{Overview of competitions in the OPC.}
    \label{fig:competitions}
\end{wrapfigure}

\paragraph{Basic properties} The OPC consists of $5{,}062$ proofs across $1{,}010$ distinct problems, generated by six state-of-the-art LLMs. Each proof is labeled as either correct or incorrect by one or two human judges. Labels are accompanied by short justifications, with optional annotations highlighting specific sentences within the proof. Each problem may also include metadata such as its competition source, difficulty level, and other relevant attributes. An example is shown in \cref{fig:overview:left}.

\paragraph{Competitions} Problems were sourced from a wide range of prestigious mathematics competitions. A full breakdown is provided in \cref{app:comp}, with a summary shown in \cref{fig:competitions}. \cref{app:comp} also reports the average accuracy of the best-performing model per competition, offering a rough proxy for difficulty. Most problems are at the high school level ($\approx 84\%$), with a small portion drawn from undergraduate-level competitions ($\approx 16\%$).

\begin{wraptable}[9]{r}{0.4\textwidth}
    \centering
    \vspace{-4mm}
    \caption{Solutions per evaluated model.}
    \vspace{-3mm}
    \label{tab:model_summary}
    \begin{tabular}{lc}
        \toprule
        {\textbf{Model}} & {\textbf{\# Solutions}}\\
        \midrule
        \textsc{o4-mini} & 1615 \\
        \textsc{o3} & 892 \\
        \textsc{Qwen3-235B-A22B} & 890 \\
        \textsc{Gemini-Pro} & 878 \\
        \textsc{Grok-3-Mini} & 461 \\
        \textsc{DeepSeek-R1} & 326 \\
        \bottomrule
    \end{tabular}
\end{wraptable}

\paragraph{Models} \cref{tab:model_summary} shows the number of proofs generated by each model. Not all models attempted every problem, but most problems include solutions from five models. \textsc{o4-mini} contributed the largest number of proofs, as it was used in the best-of-n and pass@n experiments.

\paragraph{Human performance} To estimate label reliability, we double-graded approximately $10\%$ of the dataset. Among these, judges agreed on the proof's correctness in $90.4\%$ of cases. Assuming independent judgments, we can estimate the individual judge error rate $p$ by solving $0.904 = (1 - p)^2 + p^2$, giving $p = 5\%$. This indicates a low noise level, which is expected for a human-annotated dataset of this complexity.

\paragraph{Data splits} The OPC is divided into four subsets, each serving a distinct purpose:

\begin{itemize}\setlength\itemsep{0.005em}
    \item \textbf{MathArena}: A subset of $112$ problems from MathArena \citep{matharena}, a final-answer benchmark for mathematical reasoning.
    \item \textbf{PutnamBench}: A subset of $114$ problems from PutnamBench \citep{putnambench}, used to compare natural language and formal proof generation.
    \item \textbf{Best-of-n}: A subset of $152$ problems from hard competitions, each solved multiple times by \textsc{o4-mini}. For 60 of these problems, all 8 generations were human-evaluated. The rest include judgments only for the generations selected by a best-of-n selection strategy.
    \item \textbf{Generic}: A subset of $676$ problems from various competitions, solved by multiple models.
\end{itemize}
The MathArena and PutnamBench subsets are drawn from existing benchmarks and should not be used for training. The generic and best-of-n subsets are intended for training, validation, and further analysis. However, a small portion of the generic subset is held back for benchmarking purposes.

%% file: paper_files/results.tex
\section{Results}\label{sec:results}

We now present our main findings. In \cref{sec:proofcorrectness}, we evaluate the proof-generation capabilities of various models. \cref{sec:llmjudge} evaluates the ability of LLMs to judge proof correctness. In \cref{sec:informalvsformal}, we compare informal and formal proof-generation performance. \cref{sec:finalanswercorrectness} analyzes proof correctness given correct final answers. \cref{sec:bestofn} examines the effectiveness of various best-of-n strategies. Finally, in \cref{sec:contamination}, we discuss the potential impact of data contamination on our results. In \cref{app:qualitative}, we additionally provide some qualitative observations about common mistakes in the generated proofs. 

All shown confidence intervals in this section are 95\% intervals computed using the large sample normal approximation. Bold numbers indicate the best performance in the respective category, and underlined numbers are within the confidence interval of the best performance. 

\subsection{Proof Generation Capabilities} \label{sec:proofcorrectness}
\vspace{-1mm}

\begin{figure}[t]
    \centering
    \includegraphics[width=0.65\linewidth]{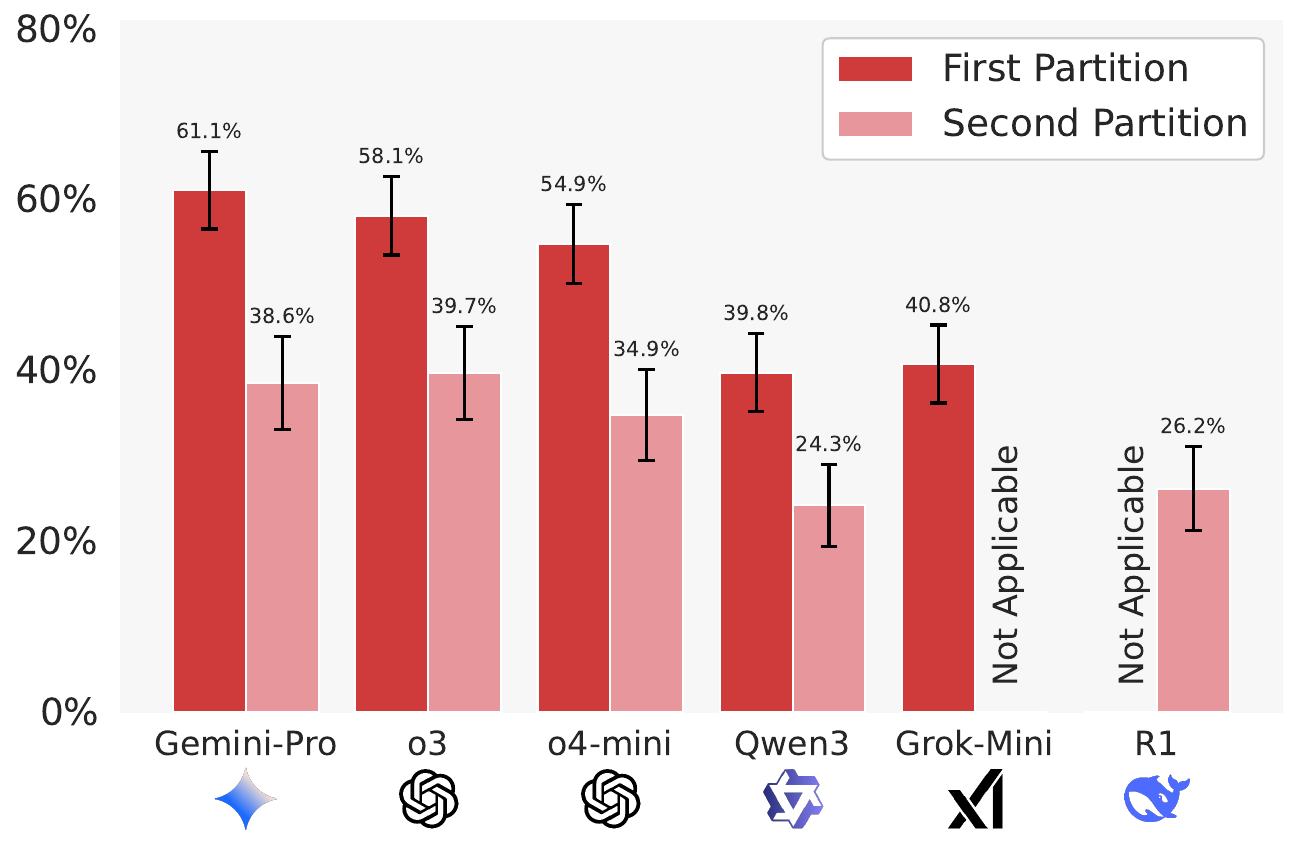}
    \vspace{-3mm}
    \caption{Average proof correctness on the OPC. Data is split into two partitions, the first, resp. second, containing only problems answered by all models except \rone{}, resp. \grokthreemini{}. The second partition contains problems from more challenging competitions, explaining the score discrepancy.}
    \vspace{-3mm}
    \label{fig:proofcorrectness}
\end{figure}

\paragraph{Proof generation results} \cref{fig:proofcorrectness} shows the average proof correctness of each model on a subset of the OPC, including only fully-judged questions for all models except \grokthreemini{} or \rone{}. \geminipro{} achieves the highest accuracy, slightly outperforming \ofour{}. In contrast, the two open-source models, \qwenthreebig{} and \rone{}, underperform significantly, highlighting the performance gap between closed- and open-source models in generating correct proofs.

\paragraph{Uncertainty acknowledgment} Out of more than 1,700 incorrect solutions analyzed, models explicitly state their inability to solve the problem in only 114 instances, with all but five of those generated by \othree{}. Even \othree{} is more likely to produce an incorrect proof than to acknowledge uncertainty. This reluctance to admit failure highlights a key limitation. In domains like mathematics, this could undermine trust in systems that rely on LLMs for provably correct solutions.

\subsection{LLMs Are Human Level Judges} \label{sec:llmjudge}
\vspace{-1mm}
\paragraph{Setup} The OPC provides binary human judgments for proof correctness, making it ideal for training and evaluating LLM proof judges. To leverage this, we split the generic subset by problem into training and test sets. Using GRPO \citep{grpo}, we fine-tune \ronesmall{} using human labels for rewards. The test set comprises 292 LLM-generated proofs with 36\% average correctness. We evaluate both reasoning models, such as \gptfive{}, and general-purpose models like \textsc{GPT-4.1}. Importantly, the human baseline is not measured on the test subset, but rather on all double-graded proofs in the OPC. Since the test samples are uniformly drawn from the OPC, this does not significantly affect the comparison.
\input{tables/judge_performance.tex}

\vspace{-1mm}
\paragraph{Results} As shown in \cref{tab:judge_summary}, \gptfive{} achieves the highest judging accuracy: $89.3\%$ with a single evaluation pass, approaching the $90.4\%$ human baseline, and $90.8\%$ with majority voting. Notably, \opcrone{} matches \geminipro{}'s majority voting performance and outperforms its base model by $17\%$, demonstrating the value of the OPC and its potential for advancing the field of proof evaluation and generation. However, the train set for \opcrone{} shares the same distribution as this test set, which may inflate its performance. In \cref{app:additional}, we show that while the performance of \opcrone{} is reduced under out-of-distribution data, the improvement over the base model persists even under these conditions.

This positive result appears to contradict \citet{prooforbluff}, who reported poor performance of judge models. However, their evaluation was based on a limited set of problems, relied on older models, and focused on the more challenging task of scoring proofs on a continuous rather than a binary scale. In addition, we put considerable effort into crafting clear and comprehensible prompts.

\input{tables/judge_prover.tex}
\paragraph{Self-evaluation} LLMs often favor their own generations \citep{favourown}. To investigate this, we evaluate how well models judge their proofs compared to those generated by others. In \cref{tab:judge_prover}, we find that all models except \qwenthreebig perform worse when judging their own proofs. This suggests that LLMs struggle to recognize their own mistakes, which is a critical limitation for applications requiring self-evaluation.

\vspace{-1mm}
\subsection{Formal Proof Generation Lags Behind} \label{sec:informalvsformal}

\begin{figure}[t]
    \centering
    \begin{minipage}[t]{0.48\textwidth}
        \centering
        \includegraphics[width=0.9\linewidth]{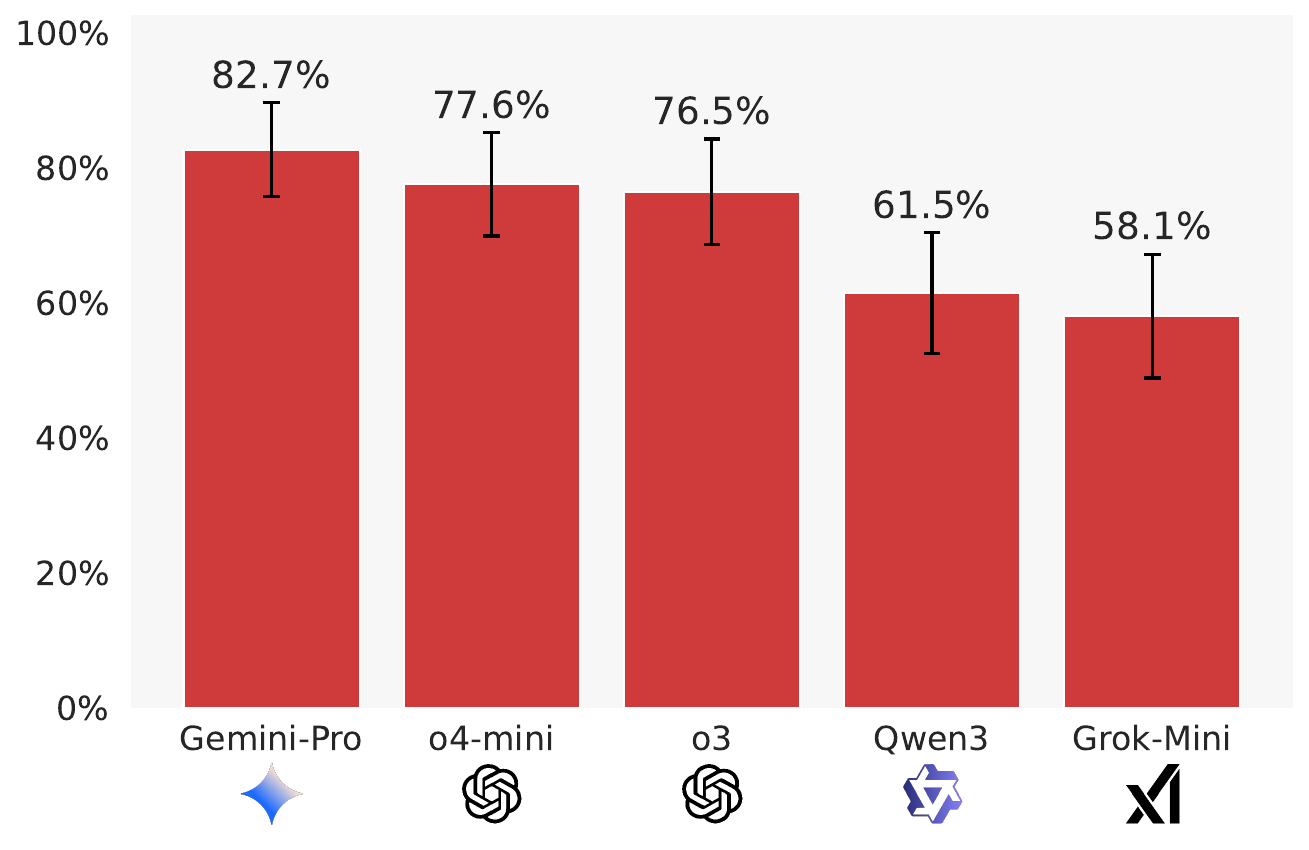}
        \vspace{-3mm}
        \caption{Average proof correctness on the PutnamBench.}
        \label{fig:putnambench}
    \end{minipage}%
    \hfill
    \begin{minipage}[t]{0.48\textwidth}
        \centering
        \includegraphics[width=0.9\linewidth]{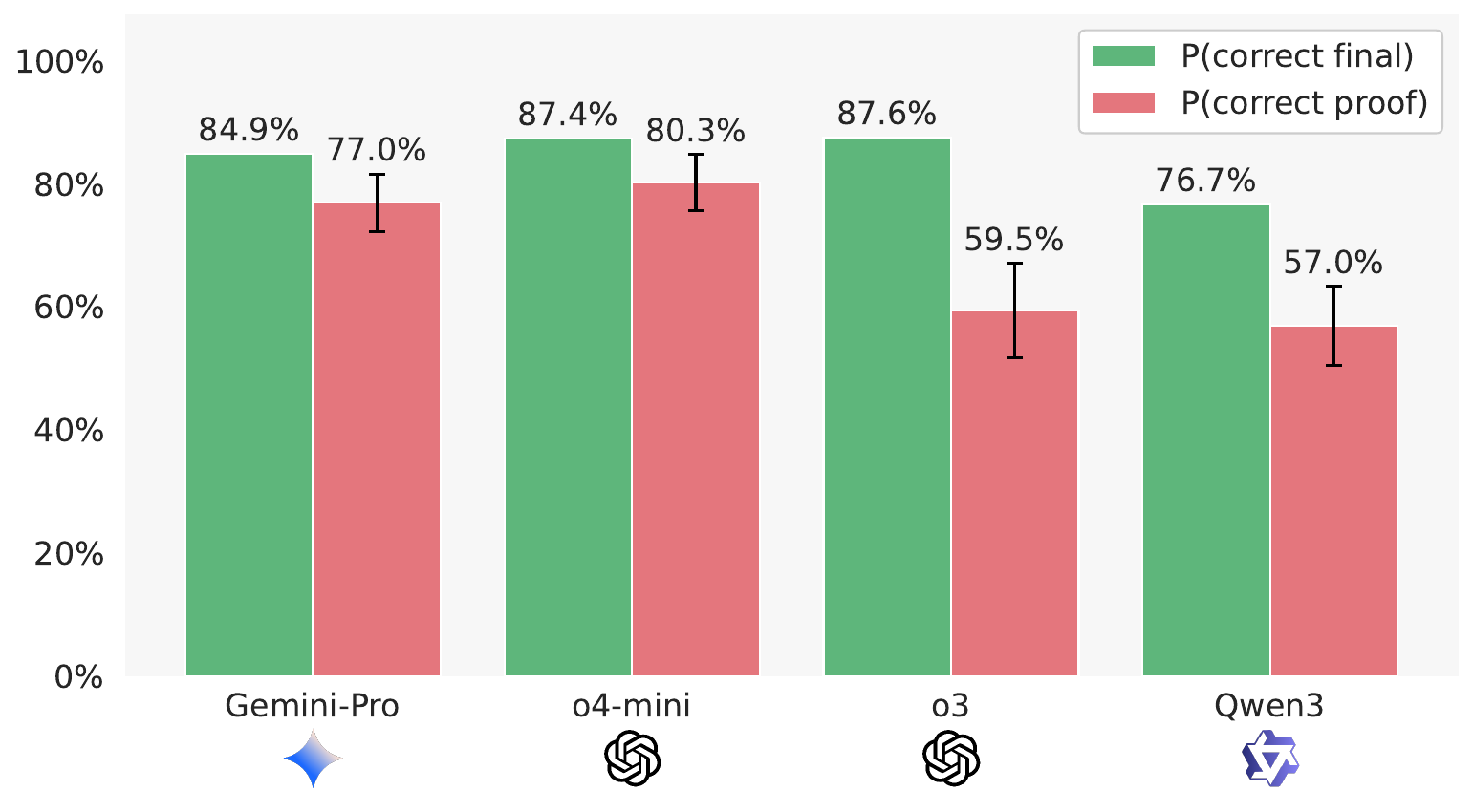}
        \vspace{-3mm}
        \caption{Comparison of final-answer accuracy and proof correctness on the MathArena subset.}
        \label{fig:matharena}
    \end{minipage}
\end{figure}
\vspace{-1mm}
Using the PutnamBench subset, we compare formal and natural language proof-generation models. The best formal model, \textsc{Goedel-Prover-v2} \citep{goedelproverv2}, achieves less than $19\%$ accuracy on the PutnamBench. In contrast, \cref{fig:putnambench} shows that the top informal model, \geminipro{}, reaches almost $83\%$ accuracy on the evaluated subset, clearly outperforming this baseline. Despite this disparity, formal proofs offer a major advantage: automatic verifiability. While informal methods currently dominate, formal approaches remain crucial for scalable proof verification. 

Finally, we note that a recent private \emph{agentic} system, Seed-Prover \citep{seedprover}, obtained a $50\%$ formal accuracy on the PutnamBench. However, our informal results do not use agentic techniques, and it is therefore not accurate to directly compare these numbers.

\subsection{Proof Generation and Final Answer Do Not Align} \label{sec:finalanswercorrectness}

Although it is widely claimed that final-answer benchmarks are inadequate for evaluating proof generation capabilities \citep{prooforbluff,brainvbytes,rstar}, it remains unclear how often LLMs can produce a valid proof when they find the correct answer. To investigate this, we first collect instances from the MathArena subset where models generate correct final answers, and then manually evaluate the validity of the accompanying proofs. This enables us to estimate the overall proof correctness rate and compare it with the final-answer accuracy.

\vspace{-1mm}
\paragraph{Results} In \cref{fig:matharena}, we report each model's final-answer accuracy alongside the stricter metric requiring a valid proof. Despite \geminipro{}, \ofour{}, and \othree{} achieving similar final-answer accuracy, proof correctness differs significantly. Specifically, \othree{} performs notably worse, with only $59.5\%$ of its answers containing a correct proof. This substantial difference between models shows that final-answer accuracy is not a reliable indicator of proof generation capability.

\subsection{Best-of-n Significantly Improves Performance} \label{sec:bestofn}

\begin{figure}[t]
    \centering
    \begin{subfigure}{0.48\linewidth}
        \centering
        \includegraphics[width=\linewidth]{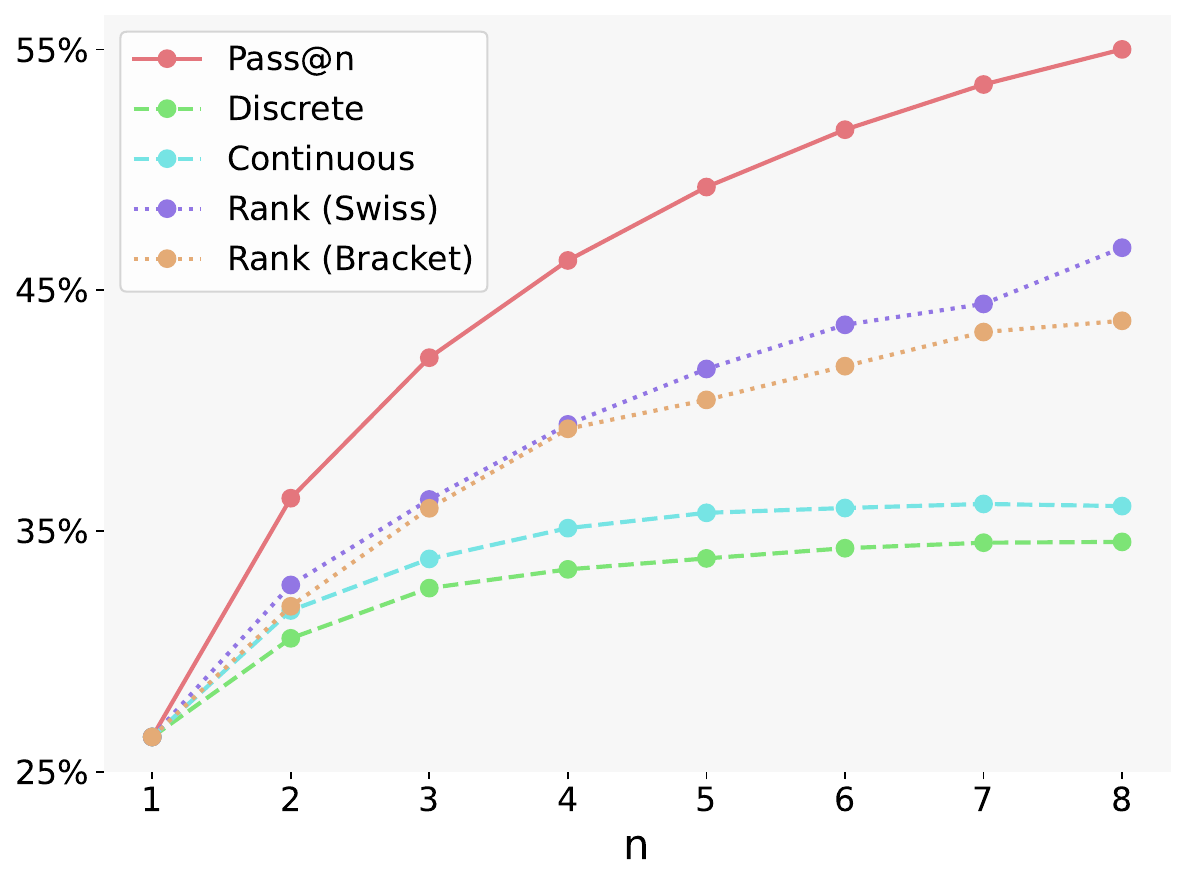}
        \vspace{-5mm}
        \caption{Pass@n versus all other best-of-n methods.}
        \label{fig:passatn}
    \end{subfigure}
    \hfill
    \begin{subfigure}{0.48\linewidth}
        \centering
        \includegraphics[width=\linewidth]{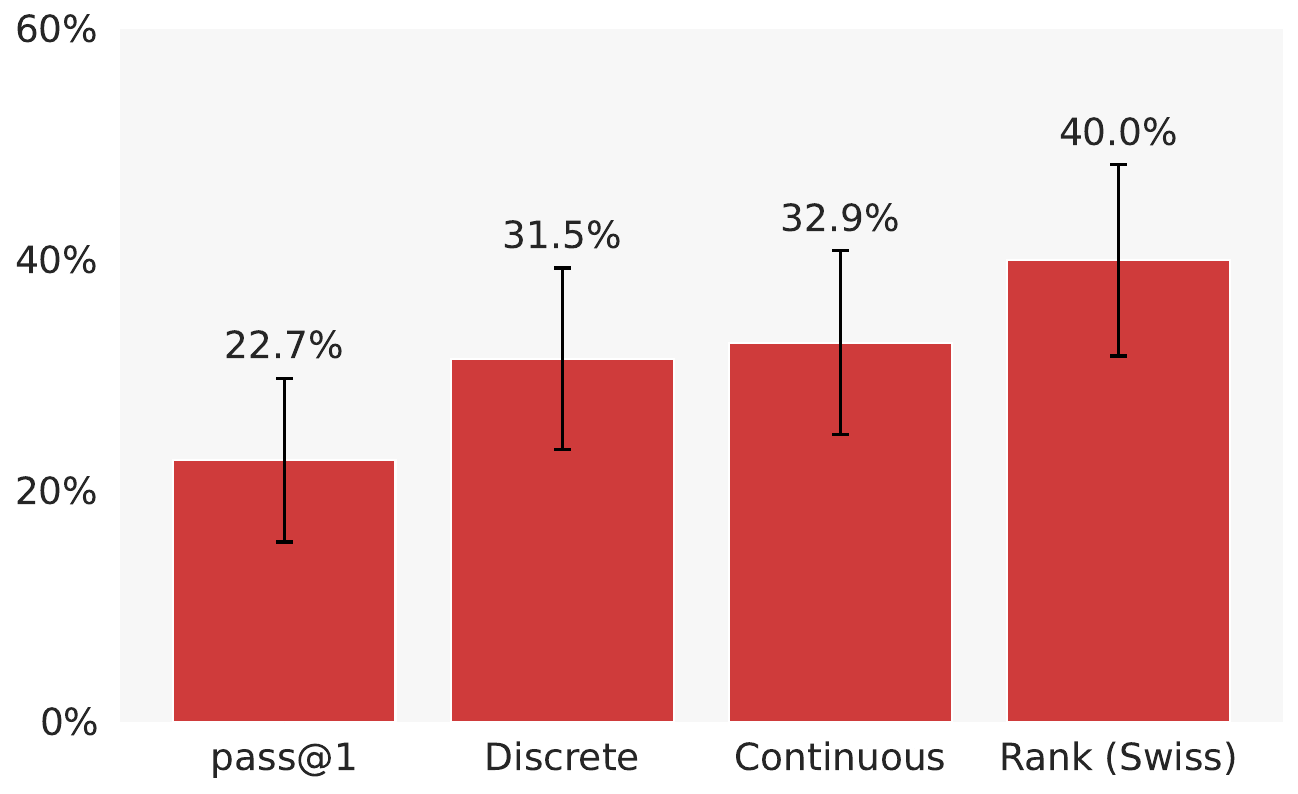}
        \vspace{-5mm}
        \caption{Selection of methods on a larger subset.}
        \label{fig:bestofn2}
    \end{subfigure}
    \vspace{-2mm}
    \caption{Performance of the best-of-n selection strategies.}
    \vspace{-3mm}
    \label{fig:comparison}
\end{figure}

\paragraph{Best-of-n selection strategies} Best-of-n selection, generating multiple outputs and selecting the best one, is a common strategy to improve performance. We evaluate this approach using \ofour{} by generating eight proofs per problem in the best-of-n subset and testing four selection methods:
\vspace{-1mm}
\begin{itemize}\setlength\itemsep{0.005em}
    \item \textbf{Discrete}: \ofour{} classifies each proof as correct or incorrect and selects any correct one.
    \item \textbf{Continuous}: \ofour{} scores proofs on a $0$-$7$ scale and selects one with the highest score.
    \item \textbf{Rank (Bracket)}: A ranking method proposed by \citet{knockout}. Proofs are judged pairwise by \ofour{} in a knockout tournament until one proof remains.%
    \item \textbf{Rank (Swiss)}: Inspired by \citet{knockout}, proofs are paired in a Swiss round-robin tournament. Ratings are computed using the Bradley-Terry model \citep{bradley1952rank}, and the proof with the highest rating is selected. See \cref{app:swiss} for details.
\end{itemize}
\vspace{-2mm}

Prompts for all methods can be found in \cref{app:prompts}. 

\vspace{-1mm}
\paragraph{Complexity} Note that \emph{Rank (Bracket)} requires $O(n)$ comparisons, making it as efficient as the discrete method, while \emph{Rank (Swiss)} requires $O(n^2)$ comparisons, making it more expensive.

\vspace{-1mm}
\paragraph{Results on small subset} In \cref{fig:passatn}, we compare the performance of these methods with the pass@n metric on the $60$ problems that contain human judgments for all eight proofs. We find that best-of-n selection strategies can significantly improve proof generation performance. Furthermore, the pairwise ranking methods outperform the discrete and continuous methods by approximately $10\%$. Notably, while discrete and continuous methods plateau after $n=5$, ranking approaches continue to scale. Finally, \emph{Rank (Swiss)} slightly outperforms \emph{Rank (Bracket)} by a $3\%$ margin.

\vspace{-1mm}
\paragraph{Results on larger subset} In \cref{fig:bestofn2}, we evaluate the performance of the best-of-n selection strategies on the $134$ problems of the best-of-n subset of the OPC\footnote{A small bug in the \emph{Rank (Swiss)} method caused incorrect selections for 18 questions. These are excluded from the analysis.}, except for the \emph{Rank (Bracket)} method, which was not evaluated on the full subset. The improved performance of the ranking methods is confirmed on the entire best-of-n subset, with \emph{Rank (Swiss)} improving accuracy by $17\%$. While the confidence intervals are relatively large due to the small number of problems in this problem set, all selection methods rely on the same underlying answers from \ofour{}, making the relative performance differences significant.

\vspace{-2mm}
\subsection{Potential Impact of Contamination} \label{sec:contamination}
\vspace{-1mm}
Since many OPC problems are publicly available, there is a possibility that some were contaminated during training. In this section, we briefly explain why this plays an insignificant role in our results.

\vspace{-2mm}
\paragraph{Proof generation capabilities} In \cref{app:additional}, we present a small experiment indicating that contamination has a smaller effect than problem difficulty. Nevertheless, it cannot be fully ruled out. In particular, the small gap in performance between \geminipro{} and \ofour{} in \cref{sec:proofcorrectness} cannot be conclusively attributed to genuine performance differences. However, results on smaller, contamination-free datasets \citep{prooforbluff,matharena} also show \geminipro{} outperforming \ofour{}, supporting the interpretation that the performance gap is real.

In all other sections, the conclusions are robust to contamination: the informal-formal gap is too large to be affected by small changes, the discrepancy between final-answer accuracy and proof correctness is based on problems from MathArena, a well-known benchmark with problems created in 2025, and best-of-n strategies are compared using the same model, making contamination irrelevant.

\input{tables/judge_performance_gt.tex}

\vspace{-2mm}
\paragraph{Proof judging capabilities} Data contamination poses a less significant risk for proof judging, since generated proofs cannot be present in the training data. The underlying problems and their official solutions, however, may have been seen during training, which could give models an advantage. To test this, we run a worst-case experiment where the ground-truth solution is provided alongside the proof to be judged. As shown in \cref{tab:judge_with_gt}, the resulting accuracy gains are small and non-significant, suggesting that access to the solution does little to improve error detection and that contamination has a limited impact on judging performance.

%% file: tables/judge_performance.tex
\begin{wraptable}[17]{r}{0.5\textwidth}
    \caption{LLMs as proof graders. Cost for running the model on the entire subset is given in USD. Full confidence intervals are given in \cref{app:results:significance}.}
    \label{tab:judge_summary}
    \vspace{-2mm} %
    \resizebox{\linewidth}{!}{

    \begin{tabular}{l
        x{2}{1}
        x{2}{1}
        x{3}{2}}
        \toprule
        {\textbf{Judge}} & {\textbf{pass@1}} & {\textbf{maj@5}} & {\textbf{Cost}} \\
        \midrule
        \textsc{Human} & 90.4 & {-} & {N/A} \\
        \gptfive{} & \bf{89.3} & \bf{90.8} & 117.77 \\
        \grokfour{} & \underline{88.3} & \underline{89.8} & 104.42 \\
        \geminipro{} & 85.4 & \underline{88.1} & 135.47 \\
        \textbf{\opcrone{}} & 83.8 & \underline{88.1} & {N/A} \\
        \ofour{} & 83.8 & 85.3 & 29.57 \\
        \othree{} & 83.1 & 84.3 & 93.3 \\
        \geminiflash{} & 82.7 & 86.0 & 86.95 \\
        \qwenthreebig{} & 81.8 & 84.6 & 3.79 \\
        \rone{} & 80.9 & 82.6 & 27.70 \\
        \ronesmall{} & 70.7 & 71.3 & {N/A} \\
        \claude{} & 70.6 & 75.0 & 28.21 \\
        \textsc{Qwen3-8B} & 64.4 & 63.6 & {N/A}  \\
        \textsc{GPT-4.1} & 61.4 & 60.8 & 20.33 \\
        \bottomrule
    \end{tabular}
    }
    \vspace{-4mm} %
\end{wraptable}

%% file: tables/judge_prover.tex
\begin{wraptable}[10]{r}{0.5\textwidth}
    \vspace{-4mm}
    \caption{Judgement accuracy breakdown, highlighting the lowest score for each judge. Full confidence intervals are given in \cref{app:results:significance}.}
    \label{tab:judge_prover}
    \resizebox{\linewidth}{!}{
        \begin{tabular}{l
            x{2}{1}
            x{2}{1}
            x{2}{1}
            x{2}{1}}
        \toprule
        
        \diagbox[width=\textwidth/6+2\tabcolsep\relax, height=0.95cm]{{Prover}}{\textbf{Judge}} &
        {\textbf{\textsc{Gemini}}} &
        {\textbf{\textsc{o4}}} &
        {\textbf{\textsc{o3}}} &
        {\textbf{\textsc{Qwen}}} \\
        \midrule
        \textsc{Gemini} & \textbf{79.4} & 86.9 & 85.9 & \underline{80.0} \\
        \textsc{o4} & 87.1 & \textbf{81.3} & 84.8 & \underline{81.9} \\
        \textsc{o3} & 91.6 & \underline{83.1} & \textbf{76.9} & \textbf{79.1} \\
        \textsc{Qwen} & \underline{80.6} & \underline{84.1} & 87.8 & 84.4 \\
        \bottomrule
        \end{tabular}
    }
\end{wraptable}

%% file: tables/judge_performance_gt.tex
\begin{wraptable}[14]{r}{0.5\textwidth}
    \vspace{-4mm} %
    \caption{Model accuracy with and without providing the ground truth solutions. $\Delta$ shows the change in accuracy between the two settings.}
    \label{tab:judge_with_gt}
    \resizebox{\linewidth}{!}{
    \begin{tabular}{lccc}
        \toprule
        Judge & pass@1 & \makecell{pass@1\\(w/ Solution)} & $\Delta$ \\
        \midrule
        \gptfive{} & 89.3 & 89.0 & -0.3 \\
        \geminipro{} & 85.4 & 82.7 & -2.7 \\
        \opcrone{} & 83.8 & 83.1 & -0.7 \\
        \ofour{} & 83.8 & 85.2 & +1.4 \\
        \othree{} & 83.1 & 85.9 & +2.8 \\
        \qwenthreebig{} & 81.8 & 84.9 & +3.1 \\
        \rone{} & 80.9 & 80.8 & -0.1  \\
        \ronesmall{} & 70.7 & 75.4 & +4.7  \\
        \bottomrule
    \end{tabular}
    }
    \vspace{-4mm} %
\end{wraptable}

%% file: paper_files/limitations.tex
\vspace{-3mm}
\section{Limitations}
\label{sec:limitations}
\vspace{-1mm}

While the OPC represents a significant advancement in the development and evaluation of LLM proof-generation capabilities, it is not without limitations. First, since dataset construction took place before the release of \grokfour{} and \gptfive{}, these models are only included as judges. However, recent benchmarks suggest that these models perform similarly, perhaps slightly better, compared to \geminipro{} \citep{matharena}. Therefore, it does not affect the validity of our conclusions. Second, the majority of problems in the OPC are derived from high-school competitions. Thus, the dataset does not cover more advanced mathematical domains, such as research-level mathematics. We further outline potential directions for expanding the OPC in \cref{sec:futurework}.

%% file: paper_files/conclusion.tex
\vspace{-1mm}
\section{Conclusion}\label{sec:conclusion}
\vspace{-1mm}
In this work, we introduced the \textit{Open Proof Corpus (OPC)}, a human-validated dataset comprising over $5{,}000$ LLM-generated proofs. Using the OPC, we addressed several open questions in the field, including (1) the gap between natural language and formal proof generation, (2) the relationship between final-answer accuracy and proof correctness, and (3) the effectiveness of best-of-n selection. We also trained a model that achieves $88.1\%$ accuracy in judging proof correctness, matching \geminipro{} and approaching the performance of the top model, \gptfive{}.

\section*{Reproducibility Statement}
We have included our dataset in the supplementary material, along with detailed descriptions of our methodology and experimental setup to ensure full reproducibility. In addition, we provide our code with step-by-step instructions in the supplementary material, enabling others to replicate our results and for finetuning a judge model. The dataset and judge model are also open-sourced in our HuggingFace repository, with clear documentation to facilitate use by the research community.

%% file: paper_files/acknowledgements.tex
\section*{Acknowledgements}
This work has received funding from the Swiss National Science Foundation (SNSF) [200021\_207967], the Ministry of Education and Science of Bulgaria (support for INSAIT, part of the Bulgarian National Roadmap for Research Infrastructure), and the Swiss State Secretariat for Education, Research and Innovation (SERI).

%% file: paper_files/appendix.tex
\input{paper_files/app_comps.tex}

\input{paper_files/app_swiss.tex}
\input{paper_files/app_additional.tex}
\input{paper_files/app_details.tex}

\input{paper_files/app_results.tex}

\input{paper_files/future_work}
\input{paper_files/app_llm.tex}
\input{paper_files/app_interface.tex}

\input{paper_files/app_prompts.tex}

%% file: paper_files/app_comps.tex
\section{Competitions in the OPC}\label{app:comp}
The OPC contains over 1000 problems that were sourced from national and international competitions of varying difficulty. In \cref{tab:comp_distribution}, we present the problem and sample distribution for each. We also include the following additional information:
\vspace{-3mm}
\begin{itemize}\setlength\itemsep{0.005em}
    \item \textbf{Level}: the education level the problems are appropriate for, either high school (HS) or undergraduate (UG).
    \item \textbf{Type}: whether the competition is hosted internationally (I) or only nationally (N).
    \item \textbf{Source}: we link the source from which we obtained the problems. Any source that is not publicly available was marked as "Private".
    \item \textbf{Acc}: the average accuracy of the best-performing model on the competition problems, which serves as a rough proxy for difficulty.
\end{itemize}

\input{tables/comp_table.tex}

%% file: tables/comp_table.tex
\begin{table}[t]
    \centering
    \caption{A list of competition sources for the problems in OPC.}
    \resizebox{\linewidth}{!}{
        \begin{tabular}{
            llcccccc
            }
            \toprule
            \textbf{Competition} & \textbf{Description} & \textbf{Problems} & {\textbf{Solutions}} & {\textbf{Level}} & {\textbf{Type}} & {\textbf{Acc}} & {\textbf{Source}} \\

\midrule
\midrule
\multicolumn{8}{c}{\textbf{Main Analysis}}\\
\midrule
\midrule
    Balkan MO Shortlist & Competition between Balkan countries & 74 & 368 & HS & I & 31.7\% & \href{https://imomath.com/srb/zadaci/}{Public} \\
    Baltic Way MO & Northern and Central European Olympiad & 80 & 395 & HS & I & 68.1\% & \href{https://www.math.olympiaadid.ut.ee/eng/html/?id=bw}{Public} \\
    British MO Final & Final round of the British Olympiad & 23 & 114 & HS & N & 78.3\% & \href{https://bmos.ukmt.org.uk/}{Public} \\
    British MO Prelim & Preliminary round for the British Olympiad & 28 & 139 & HS & N & 87.5\% & \href{https://bmos.ukmt.org.uk/}{Public} \\
    Bulgarian Seasonal Competitions & Seasonal Competitions hosted in Bulgaria (8th-12th grade) & 49 & 242 & HS & N & 63.5\% & \href{https://klasirane.com/}{Public} \\
    European Girls' MO & Europe-wide olympiad allowing only girls as participants & 47 & 227 & HS & I & 37.8\% & \href{https://web.evanchen.cc/problems.html}{Public} \\
    IMC & International competition for university students & 43 & 212 & UG & I & 62.2\% & \href{https://www.imc-math.org.uk/}{Public} \\
    IMO Shortlist & Shortlist of problems, from which the IMO is selected & 128 & 629 & HS & I & 18.0\% & \href{https://www.imo-official.org/}{Public} \\
    International Zhautykov MO & Kazakhstan-based olympiad with near-IMO-level questions & 41 & 203 & HS & I & 34.2\% & \href{https://izho.kz/contest/problems/}{Public} \\
    Irish MO & Final round of the Irish Olympiad & 51 & 255 & HS & N & 92.0\% & \href{https://bmos.ukmt.org.uk/}{Public} \\
    Putnam & Undergraduate competition, regarded as one of the most difficult & 8 & 35 & UG & I & 100.0\% & \href{https://kskedlaya.org/putnam-archive/}{Public} \\
    Romanian Masters of Mathematics Extralist & IMO-level competition hosted in Romania & 33 & 162 & HS & I & 43.3\% & Private \\
    Swiss MO & Various problems from the Swiss Olympiad & 8 & 39 & HS & N & 87.5\% & \href{https://mathematical.olympiad.ch/en/}{Public} \\
    USA Junior MO & USA olympiad that allows only junior students & 25 & 121 & HS & N & 52.2\% & \href{https://web.evanchen.cc/problems.html}{Public} \\
    USAMO & The final round of the USA Math Olympiad & 38 & 190 & HS & N & 35.1\% & \href{https://web.evanchen.cc/problems.html}{Public} \\
\midrule
\midrule
\multicolumn{8}{c}{\textbf{PutnamBench}}\\
\midrule
\midrule
    Putnam & Undergraduate competition, regarded as one of the most difficult & 114 & 564 & UG & I & 82.7\% & \href{https://raw.githubusercontent.com/trishullab/PutnamBench/refs/heads/main/informal/putnam.json}{Public} \\
\midrule
\midrule
\multicolumn{8}{c}{\textbf{MathArena}}\\
\midrule
\midrule
    AIME 2025 & Answer-based competition, serving as a qualifier for the USAMO & 24 & 93 & HS & N & 95.7\% & \href{https://matharena.ai/}{Public} \\
    BRUMO 2025 & Answer-based competition hosted by Brown University & 28 & 114 & HS & N & 100.0\% & \href{https://matharena.ai/}{Public} \\
    HMMT February 2025 & Answer-based competition hosted by Harvard and MIT & 26 & 103 & HS & N & 97.8\% & \href{https://matharena.ai/}{Public} \\
    SMT 2025 & Answer-based competition hosted by Stanford & 34 & 128 & HS & N & 92.6\% & Private \\
\midrule
\midrule
\multicolumn{8}{c}{\textbf{Best-of-n}}\\
\midrule
\midrule
    Balkan MO Shortlist & Competition between Balkan countries & 45 & 287 & HS & I & 62.5\% & \href{https://imomath.com/srb/zadaci/}{Public} \\
    IMO Shortlist & Shortlist of problems, from which the IMO is selected & 57 & 269 & HS & I & 30.8\% & \href{https://www.imo-official.org/}{Public} \\
    USAMO & The final round of the USA Math Olympiad & 40 & 173 & HS & N & 66.7\% & \href{https://web.evanchen.cc/problems.html}{Public} \\

\bottomrule
        \end{tabular}
    }
    \label{tab:comp_distribution}
\end{table}

%% file: paper_files/app_swiss.tex
\section{Swiss Ranking Methodology}\label{app:swiss}

We briefly describe the Swiss ranking method used as a best-of-n selection strategy. In this approach, a round-robin tournament is performed where each proof competes against every other. In each "game", two proofs are compared by \ofour{}, which decides which proof is better, or if they are equally good.

To determine the overall winner, we compute a rating for each proof using the Bradley-Terry model \citep{bradley1952rank}, a probabilistic model for paired comparisons commonly applied in LLM evaluation \citep{llmasajudge,polyrating}. The Bradley-Terry model estimates the probability that a proof with rating $r_i$ beats a proof with rating $r_j$ as:

\begin{equation*}
P(i \text{ beats } j) = \frac{1}{1 + \exp(r_j - r_i)}.
\end{equation*}

We fit the model to the outcomes of the round-robin tournament using maximum likelihood estimation, resulting in a rating for each proof. The proof with the highest rating is selected as the best.

%% file: paper_files/app_additional.tex
\section{Additional Results}\label{app:additional}

In this section, we present further results, specifically addressing critical aspects of our evaluation: out-of-distribution performance of our trained model and the potential impact of data contamination.

\vspace{-1mm}
\subsection{\opcrone{} performance on OOD problems}
\vspace{-1mm}
During the training of \opcrone{}, our dataset primarily consisted of samples from high-school level competitions, aligning with the distribution of the test set. Crucially, undergraduate-level problems were explicitly excluded from training.\footnote{Due to a small bug, not all undergraduate-level problems were excluded. $5\%$ of the training data contains undergraduate problems, with $0.7\%$ from the Putnam competition. Due to the small percentage, we still consider undergraduate-level problems mostly out-of-distribution.} To evaluate \opcrone{}'s generalization capabilities on out-of-distribution data, we evaluated its judging performance on 560 undergraduate-level solution samples sourced from the Putnam competition.

\input{tables/judge_performance_undergrad.tex}

As shown in \cref{tab:judge_putnam}, \opcrone{} retains a notable improvement over its base model. While frontier models like \ofour{} achieve higher performance, \opcrone{} is nevertheless competitive, performing on par with \geminipro{}. This shows that the OPC data can be applied to mathematical domains beyond its scope.

\vspace{-1mm}
\subsection{Problem Category and Difficulty Affect Incorrect Proof Rates}
\begin{figure}[t]
    \centering
    \begin{subfigure}{0.48\linewidth}
        \centering
        \includegraphics[width=\linewidth]{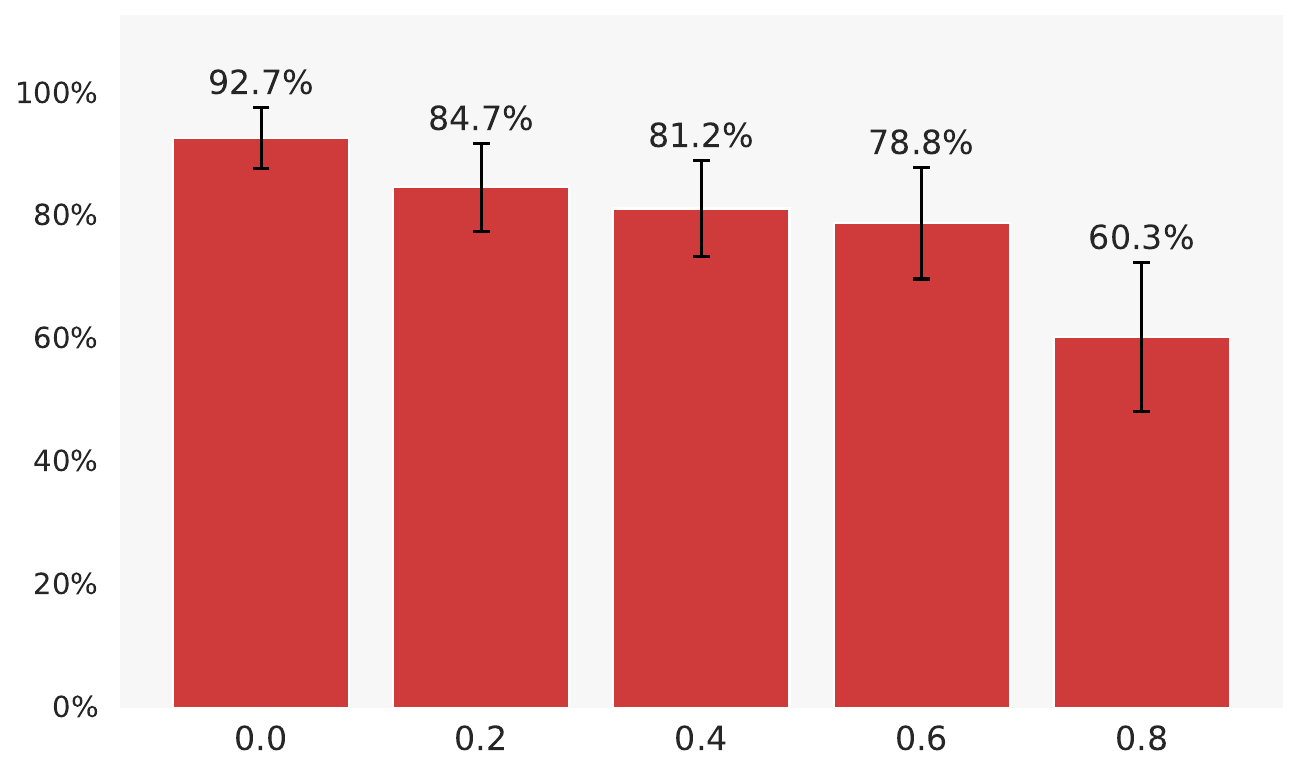}
        \vspace{-5mm}
        \caption{Problems are binned by difficulty level, with higher levels indicating more challenging problems.}
        \label{fig:difficulty}
    \end{subfigure}
    \hfill
    \begin{subfigure}{0.48\linewidth}
        \centering
        \includegraphics[width=\linewidth]{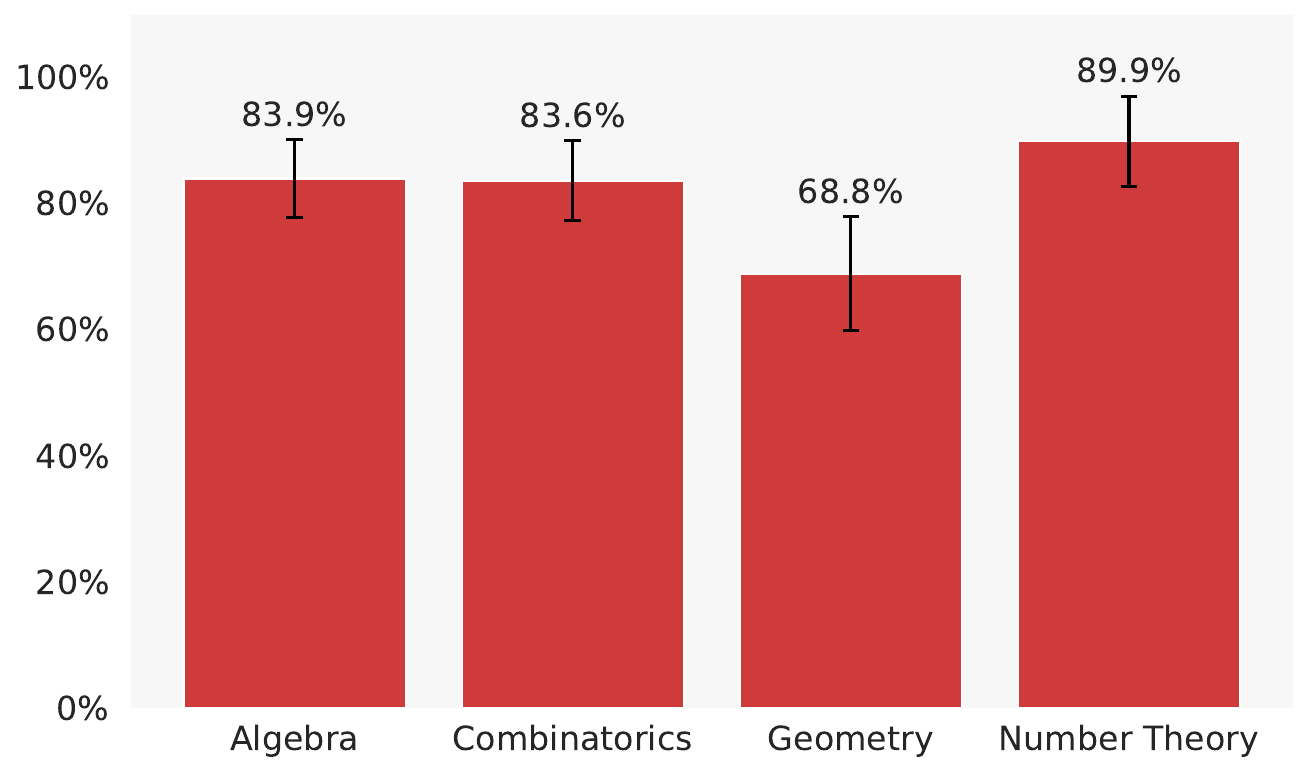}
        \vspace{-5mm}
        \caption{Problems are binned in the four main categories: Algebra, Combinatorics, Geometry, and Number Theory.}
        \label{fig:categories}
    \end{subfigure}
    \vspace{-2mm}
    \caption{Influence of difficulty level and problem category on the prevalence of incorrect proofs in MathArena.}
    \vspace{-3mm}
    \label{fig:incorrect_proofs}
\end{figure}

\vspace{-1mm}
We analyze the prevalence of incorrect proofs in the MathArena subset of the OPC. In particular, we investigate whether difficulty and problem type correlates with the likelihood of a model generating an incorrect proof. As shown in \cref{fig:incorrect_proofs}, we find that both these factors significantly influence performance. In particular, higher difficulty levels correspond to a greater proportion of incorrect proofs, as illustrated in \cref{fig:difficulty}. Additionally, among the four main problem categories, Geometry problems exhibit the highest rate of incorrect proofs.

\vspace{-1mm}

\subsection{Data Contamination for Mathematical Proofs}
\vspace{-1mm}
Test-set contamination is a prevalent challenge in evaluating the true reasoning capabilities of models, particularly in mathematics \citep{matharena, gsm1k}. Given that the vast majority of OPC problems are publicly sourced, we conduct one additional experiment to address potential contamination-related concerns.

\vspace{-1mm}
\paragraph{Competition-based breakdown}
First, we outline 2 categories of competitions: \emph{Standard} and \emph{Non-standard}. The former includes important competitions in the mathematics community, namely the IMO Shortlist and USAMO. Problems from these widely recognized competitions are more likely to appear in training sets, as they are widely reported on the internet and therefore represent the greatest potential risk for contamination.

\begin{figure}[t]
    \centering
    \includegraphics[width=0.65\linewidth]{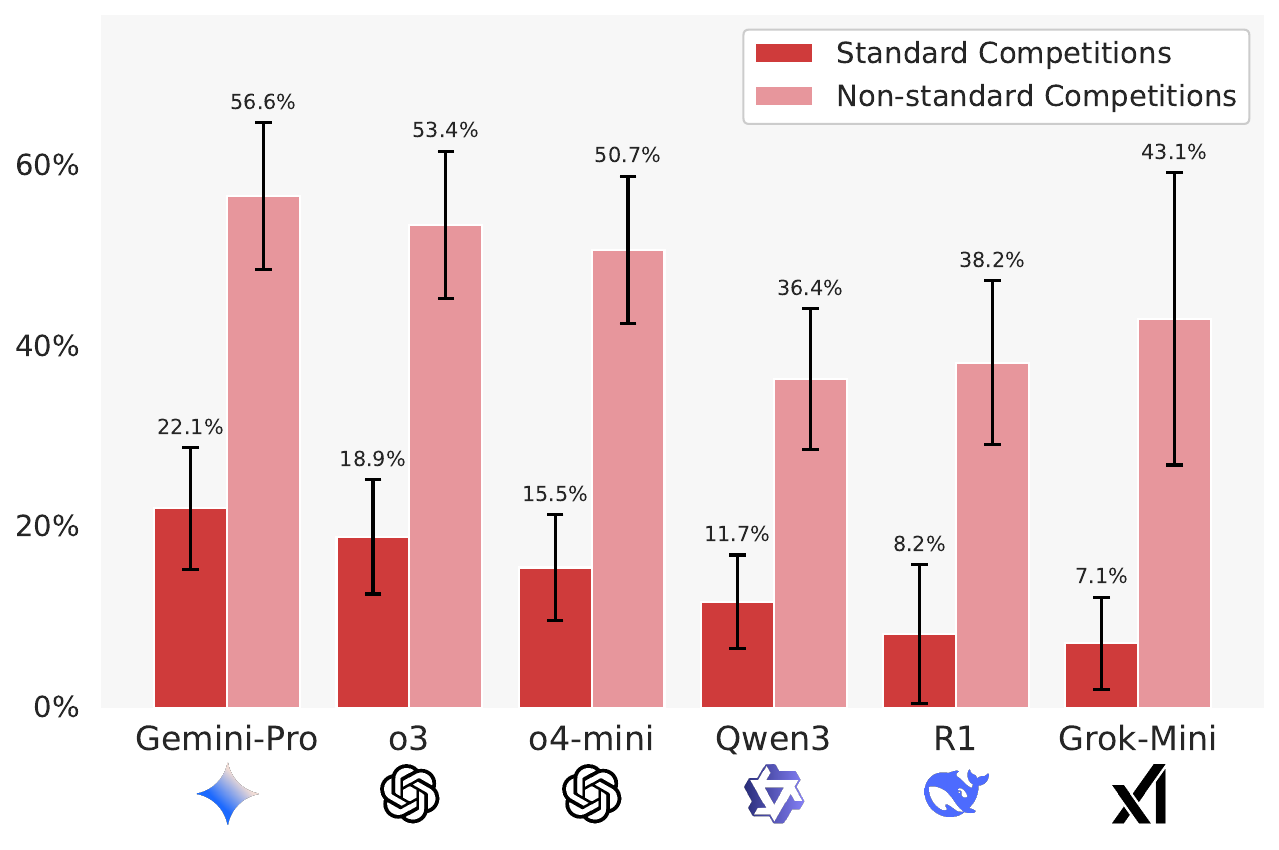}
    \vspace{-3mm}
    \caption{Average proof correctness on a sample of the OPC, split into \emph{Standard} and \emph{Non-standard} competitions.}
    \vspace{-3mm}
    \label{fig:proofcorrectness_contamination}
\end{figure}

On the other hand, the \emph{Non-standard} category consists of problems from the International Zhautykov Olympiad, the Swiss Mathematical Olympiad, and the Bulgarian seasonal competitions. A significant portion of these problems required translation from their original non-English statements, and they originate from less prevalent sources compared to the IMO and USAMO. Consequently, we anticipate a substantially lower prevalence of contamination within this problem set.

As illustrated in \cref{fig:proofcorrectness_contamination}, model accuracy on \emph{Standard} problems is notably lower than on their \emph{Non-standard} counterparts. This observation indicates that problem difficulty is a considerably more dominant factor influencing model performance than the potential presence of test-set contamination. While this experiment alone cannot definitively discount the presence of test-set contamination within the OPC dataset for our \cref{sec:proofcorrectness} results, it demonstrates that the models' problem-solving capabilities extend significantly beyond memorization. This suggests that any contamination effect is likely small and not the primary driver of performance differences.

\subsection{Significance Analysis} \label{app:results:significance}

In \cref{tab:judge_summary:significance}, we present the performance of various LLMs as proof judges, along with 95\% confidence intervals using the large sample normal approximation. In \cref{tab:judge_prover:significance}, we provide a detailed breakdown of judging accuracy by prover, again with 95\% confidence intervals. These tables complement the main results presented in \cref{sec:results}.

\begin{table}[t]
    \centering
    \caption{LLMs as proof graders. Cost for running the model on the entire subset is given in USD. Confidence intervals are computed using the large sample normal approximation.}
    \label{tab:judge_summary:significance}
    \vspace{-2mm} %

    \begin{tabular}{l
        c
        c
        x{3}{2}}
        \toprule
        {\textbf{Judge}} & {\textbf{pass@1}} & {\textbf{maj@5}} & {\textbf{Cost}} \\
        \midrule
        \textsc{Human} & $90.4 \pm 2.6$ & {-} & {N/A} \\
        \gptfive{} & $89.3 \pm 3.5$ & $90.8 \pm 3.3$ & 117.77 \\
        \grokfour{} & $88.3 \pm 3.7$ & $89.8 \pm 3.5$ & 104.42 \\
        \geminipro{} & $85.4 \pm 4.0$ & $88.1 \pm 3.7$ & 135.47 \\
        \textbf{\opcrone{}} & $83.8 \pm 4.2$ & $88.1 \pm 3.7$ & {N/A} \\
        \ofour{} & $83.8 \pm 4.2$ & $85.3 \pm 4.1$ & 29.57 \\
        \othree{} & $83.1 \pm 4.3$ & $84.3 \pm 4.2$ & 93.3 \\
        \geminiflash{} & $82.7 \pm 4.3$ & $86.0 \pm 4.0$ & 86.95 \\
        \qwenthreebig{} & $81.8 \pm 4.4$ & $84.6 \pm 4.1$ & 3.79 \\
        \rone{} & $80.9 \pm 4.5$ & $82.6 \pm 4.3$ & 27.70 \\
        \ronesmall{} & $70.7 \pm 5.2$ & $71.3 \pm 5.2$ & {N/A} \\
        \claude{} & $70.6 \pm 5.2$ & $75.0 \pm 5.0$ & 28.21 \\
        \textsc{Qwen3-8B} & $64.4 \pm 5.5$ & $63.6 \pm 5.5$ & {N/A}  \\
        \textsc{GPT-4.1} & $61.4 \pm 5.6$ & $60.8 \pm 5.6$ & 20.33 \\
        \bottomrule
    \end{tabular}
\end{table}

\begin{table}[t]
\centering
\caption{Judgement accuracy breakdown by prover with 95\% confidence intervals.}
\label{tab:judge_prover:significance}
\begin{tabular}{l
    cccc}
\toprule
        \diagbox[width=\textwidth/6+2\tabcolsep\relax, height=0.95cm]{{Prover}}{\textbf{Judge}} &
{\textbf{Gemini}} &
{\textbf{o4}} &
{\textbf{o3}} &
{\textbf{Qwen}} \\
\midrule
\textsc{Gemini} & \textbf{$79.4 \pm 4.9$} & $86.9 \pm 4.1$ & $85.9 \pm 4.2$ & $80.0 \pm 4.9$ \\
\textsc{o4} & $87.1 \pm 4.1$ & \textbf{$81.3 \pm 4.8$} & $84.8 \pm 4.4$ & $81.9 \pm 4.7$ \\
\textsc{o3} & $91.6 \pm 3.4$ & $83.1 \pm 4.6$ & \textbf{$76.9 \pm 5.2$} & \textbf{$79.1 \pm 5.0$} \\
\textsc{Qwen} & $80.6 \pm 4.9$ & $84.1 \pm 4.6$ & $87.8 \pm 4.1$ & $84.4 \pm 4.5$ \\
\bottomrule
\end{tabular}
\end{table}

%% file: tables/judge_performance_undergrad.tex
\begin{table}[t]
    \centering
    \caption{LLMs as proof graders on undergraduate-level problems. Cost for running the model on the entire subset is given in USD. Confidence intervals are 95\% and computed using the large sample normal approximation.}
    \label{tab:judge_putnam}

    \begin{tabular}{lccc}
        \toprule
        Judge & pass@1 & maj@5 & Cost \\
        \midrule
        \ofour{} & $83.8 \pm 3.1$ & $85.1 \pm 2.9$ & 48.31 \\
        \gptfive{} & $83.6 \pm 3.1$ & $84.0 \pm 3.0$ & 89.42 \\
        \rone{} & $78.3 \pm 3.4$ & $79.7 \pm 3.3$ & 51.86 \\
        \opcrone{} & $75.0 \pm 3.6$ & $77.0 \pm 3.5$ & N/A \\
        \ronesmall{} & $72.1 \pm 3.7$ & $72.4 \pm 3.7$ & N/A \\
        \geminipro{} & $70.9 \pm 3.8$ & $76.7 \pm 3.5$ & 160.95 \\
        \bottomrule
    \end{tabular}
\end{table}

%% file: paper_files/app_details.tex
\section{Experimental Details} \label{app:details}

In this section, we provide additional details regarding our experimental setup, including model training, hyperparameters, and evaluation protocols.

\subsection{Model Training}\label{app:details:model_training}
\paragraph{Training data} As outlined in \cref{sec:results}, we split the generic split of the OPC dataset into a train and test set. The training set consists of 1,733 proof samples, while the test set contains 292 proof samples. Importantly, we ensure that no problem statements overlap between the training and test sets, maintaining the integrity of our evaluation.

\paragraph{Reinforcement learning} We fine-tune \ronesmall{} using GRPO \citep{grpo} on the training set using the popular VERL framework \citep{verl} with a learning rate of $10^{-6}$, a maximum response length of $14000$ tokens, $10$ rollouts per problem, and a batch size of $16$. We use the same prompt for training as for evaluation, as shown in \cref{app:llm_as_judge_prompt}. More hyperparameters and training details can be found in the provided code. Importantly, we did not use the test set at any point during training or hyperparameter tuning, and only evaluated the final trained model once on the test set.

\subsection{Model Evaluation}\label{app:details:model_evaluation}
\paragraph{Hyperparameters and prompt} For all models, we use the recommended hyperparameters outlined by the model providers. For open models, this usually involves using nucleus sampling with $p=0.95$ and a temperature of $0.6$. For closed models, we use the default settings provided by the API. Model-specific hyperparameters can be found in the provided code. Prompts used for evaluation are shown in \cref{app:prompts}.

\paragraph{Formal model performance} To accurately compute the performance of \textsc{Goedel-Prover-V2} \citep{goedelproverv2} on our subset of problems from \citet{putnambench}, we contacted the authors of the models to obtain the raw results for each problem. This allowed us to compute the exact accuracy on our selected problems, which was not reported in the original paper.

%% file: paper_files/app_results.tex
\clearpage
\section{Qualitative Observations}\label{app:qualitative}

Building on our quantitative analysis, we present several qualitative insights drawn from our judges' evaluations. These include common failure modes and notable model behaviors.

\paragraph{Incorrect handling of inequalities}
Many models sometimes make mistakes when manipulating inequalities. A frequent issue is the improper reversal of inequality signs during intermediate steps without mathematical justification. This results in incorrect final answers despite seemingly coherent reasoning. Below is an example of such a mistake made by \rone{}, where the incorrect sign reversal is marked with "**".

\begin{deepseekls}{\deepseekheader -- Baltic Way 2004 P3}
...

Summing these inequalities gives

$
\sum_{\text{cyc}} \frac{a + b}{a + b + 1} \geq \sum_{\text{cyc}} \left( 1 - \frac{1}{3 (ab)^{1/3}} \right) = 3 - \frac{1}{3} \sum_{\text{cyc}} (ab)^{-1/3}.
$

...

By AM-GM again,

$
a^{1/3} + b^{1/3} + c^{1/3} \geq 3 \sqrt[3]{a^{1/3} b^{1/3} c^{1/3}} = 3 (abc)^{1/9} = 3 \cdot 1^{1/9} = 3,
$

with equality if and only if $a^{1/3} = b^{1/3} = c^{1/3}$, i.e., $a = b = c$. Thus,

$
\sum_{\text{cyc}} \frac{a + b}{a + b + 1} **\geq** 3 - \frac{1}{3} \cdot 3 = 3 - 1 = 2.
$
...
\end{deepseekls}

\paragraph{Struggles with synthetic geometry reasoning}
In Olympiad-level geometry problems, models often default to coordinate-based methods, i.e., using computations with complex numbers or a specific coordinate system to "bash" the problem, especially on easier problems where such techniques are tractable. However, for harder problems, models tend to shift toward synthetic approaches, relying on geometric theorems and properties. These harder problems typically require keeping track of multiple different but related angles and a robust understanding of geometric relationships. Here, models frequently struggle: they sometimes confuse angle orientations, misidentify congruent angles, reverse point orderings, or apply geometric properties incorrectly. In the following example, \geminipro{} correctly identifies the base of an isosceles triangle in one case, but subsequently deduces the wrong pair of equal angles—illustrating a common angle-reasoning failure.

\begin{geminils}{\geminiproheader -- IMOSL 2013 G4}
...
Thus, $\angle PBA = \angle QBA$ where these angles are $\angle(BP,BA)$ and $\angle(AB,BQ)$. The problem states these are equal to $\gamma$.

So, $\angle(BP,BA) = \gamma$ and $\angle(AB,BQ) = \gamma$. This means
$
\angle PBQ = \angle(BP,BA) + \angle(AB,BQ) = \gamma + \gamma = 2 \gamma.
$

We are given that $D$ is an interior point of segment $BQ$ and $PD=PB$.

Thus, $\triangle PBD$ is an isosceles triangle with base $BD$.

The angle $\angle PBD$ in $\triangle PBD$ is $\angle PBQ$, so $\angle PBD = 2\gamma$.

The base angles are
$
\angle BPD = \angle BDP = \frac{180^\circ - 2\gamma}{2} = 90^\circ - \gamma.
$
So, $\angle PDB = 90^\circ - \gamma$. Since $D$ is on segment $BQ$, this is $\angle(DP, DB) = 90^\circ - \gamma$.
...
\end{geminils}

\paragraph{Incorrect citation of IMOSL 2004 G8 by \geminipro{}} 
A curious behavior observed exclusively in \textsc{Gemini-Pro} is the repeated and inappropriate citation of the 2004 IMO Shortlist geometry problem G8. This reference appears in over 15 different solutions, often in problems unrelated to geometry. Notably, these citations consistently occur at points where the model appears to get stuck. No other model in our evaluation displayed this behavior. An illustrative example is shown below.

\begin{problem}{EGMO 2018 P3}
The $ n $ contestants of EGMO are named $ C_{1}, C_{2}, \ldots, C_{n} $. After the competition, they queue in front of the restaurant according to the following rules.
    
    - The Jury chooses the initial order of the contestants in the queue.
    - Every minute, the Jury chooses an integer $ i $ with $ 1 \leq i \leq n $.
    - If contestant $ C_{i} $ has at least $ i $ other contestants in front of her, she pays one euro to the Jury and moves forward in the queue by exactly $ i $ positions.
    - If contestant $ C_{i} $ has fewer than $ i $ other contestants in front of her, the restaurant opens and the process ends.
    
For every $ n $, prove that this process must terminate and determine the maximum number of euros that the Jury can collect by cunningly choosing the initial order and the sequence of moves.
\end{problem}
\begin{geminils}{\geminiproheader}
The problem asks us to prove that the process always terminates and to find the maximum number of euros the Jury can collect. One euro is collected for each move a contestant makes.
    
...
So we have two relations:
    $k \cdot m_k = j_k$ for each $k=1, \ldots, n$ (positional balance for each $C_k$).
    $\sum_{k=1}^n k \cdot m_k = \sum_{k=1}^n j_k$ (total jump events balance).

Substitute (1) into (2):
$
\sum_{k=1}^n k \cdot m_k = \sum_{k=1}^n (k \cdot m_k).
$
This equation is an identity and does not help by itself.

Let us use the argument from the ISL 2004 G8 solution notes:

The sum of positions is $P = \sum_{k=1}^n p_k = n(n+1)/2$, which is constant.

The sum of costs, $C = \sum_{k=1}^n (\text{position of } C_k) \times k$.
...
\end{geminils}

\paragraph{Overgeneralization from simple cases}
Consistent with prior work \citep{prooforbluff, brainvbytes}, we observe that models often overgeneralize patterns from small or simple examples. This tendency leads to incorrect solutions, particularly when the identified properties do not extend to the general case. An example illustrating this pattern overreach is included below.
 
\begin{groksl}{\grokheader -- IMO SL 2000 G4}
Let $ A_1A_2A_3 $ be an acute-angled triangle. Denote the foot of the altitude from vertex $ A_i $ to the side opposite $ A_i $ as $ K_i $, and the point where the incircle touches the side opposite $ A_i $ as $ L_i $. The lines $ K_1K_2 $, $ K_2K_3 $, and $ K_3K_1 $ are the sides of the orthic triangle, and the lines $ L_1L_2 $, $ L_2L_3 $, and $ L_3L_1 $ are the sides of the contact triangle.

...

In the special case of an equilateral triangle, the orthic triangle and contact triangle coincide, and the reflections result in the same lines, with vertices on the incircle. For a general acute-angled triangle, coordinate geometry calculations in specific examples (e.g., a triangle with vertices at $ B(0,0) $, $ C(4,0) $, $ A(1,2) $) confirm that the intersection points lie on the incircle, with numerical discrepancies attributable to rounding errors.

Given the symmetry of the configuration and the properties of the incircle as the unique circle tangent to the sides, the intersection points $ P $, $ Q $, and $ R $ lie on the incircle for any acute-angled triangle. Thus, the triangle formed by $ M_{12} $, $ M_{23} $, and $ M_{31} $ has vertices on the incircle.
\end{groksl}

%% file: paper_files/future_work.tex
\section{Future Work}\label{sec:futurework}
The OPC offers a valuable resource for advancing research in proof generation with LLMs, supporting both the training and evaluation of models for this task. Our analysis also highlights several critical gaps and challenges that require further research. First, the significant disparity between formal and natural language proof generation shows the need for more effective strategies to bridge this gap. Second, the OPC shows that current benchmarks fail to capture the full complexity and quality of generated proofs as they often rely solely on final answers. Therefore, the development of a scalable benchmarking pipeline tailored to proof generation tasks is necessary. Finally, while our results show that best-of-n sampling strategies can meaningfully improve proof quality, further research is required to better understand and optimize these methods.

%% file: paper_files/app_llm.tex
\section{Statement for the Use of Large Language Models } 

Beyond being the subject of our research, LLMs were used only as assistants to improve the clarity and quality of writing. They were not involved in aiding our research methodology, ideation, or for discovering related work.

%% file: paper_files/app_interface.tex
\section{Grading Interface and Instructions} \label{app:grading_interface}

This appendix outlines the grading interface and the accompanying instructions provided to judges. The full interface and documentation can be accessed and reviewed in our supplementary material.

\paragraph{Judge ID}  Each judge received a unique identifier, which served as their login credential on our website. This ID was used to track grading progress while maintaining judge anonymity in the resulting dataset. To facilitate discussion and resolve ambiguities, a shared communication channel was created between all judges.

\begin{figure}[t]
    \centering
    \includegraphics[width=0.8\textwidth]{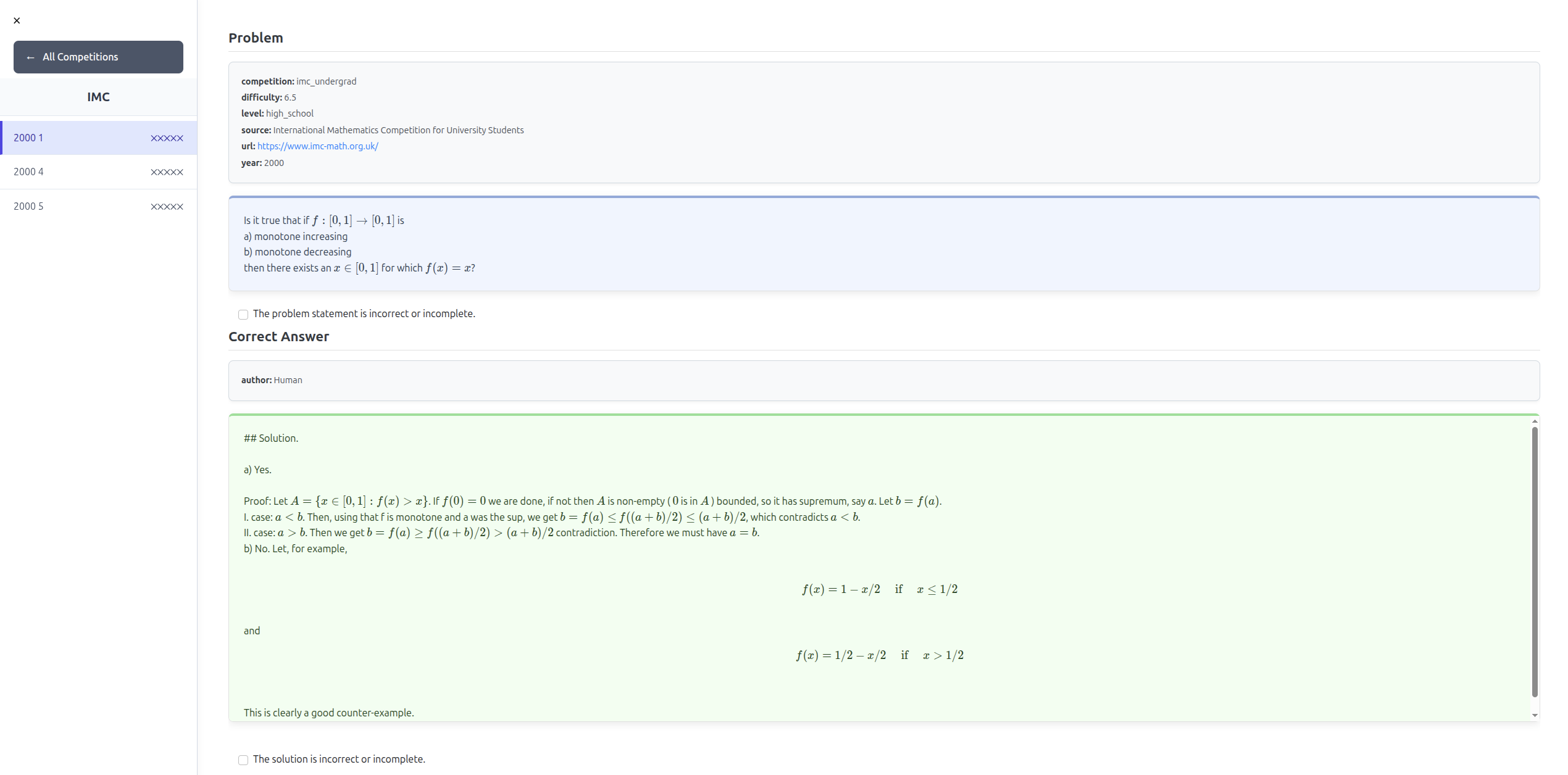}
    \caption{Main grading interface. The left panel provides navigation across problems and competitions. The right panel displays the problem and ground-truth solution, with options to report issues.}
    \label{fig:grading_interface_1}
\end{figure}

\begin{figure}[t]
    \centering
    \includegraphics[width=0.8\textwidth]{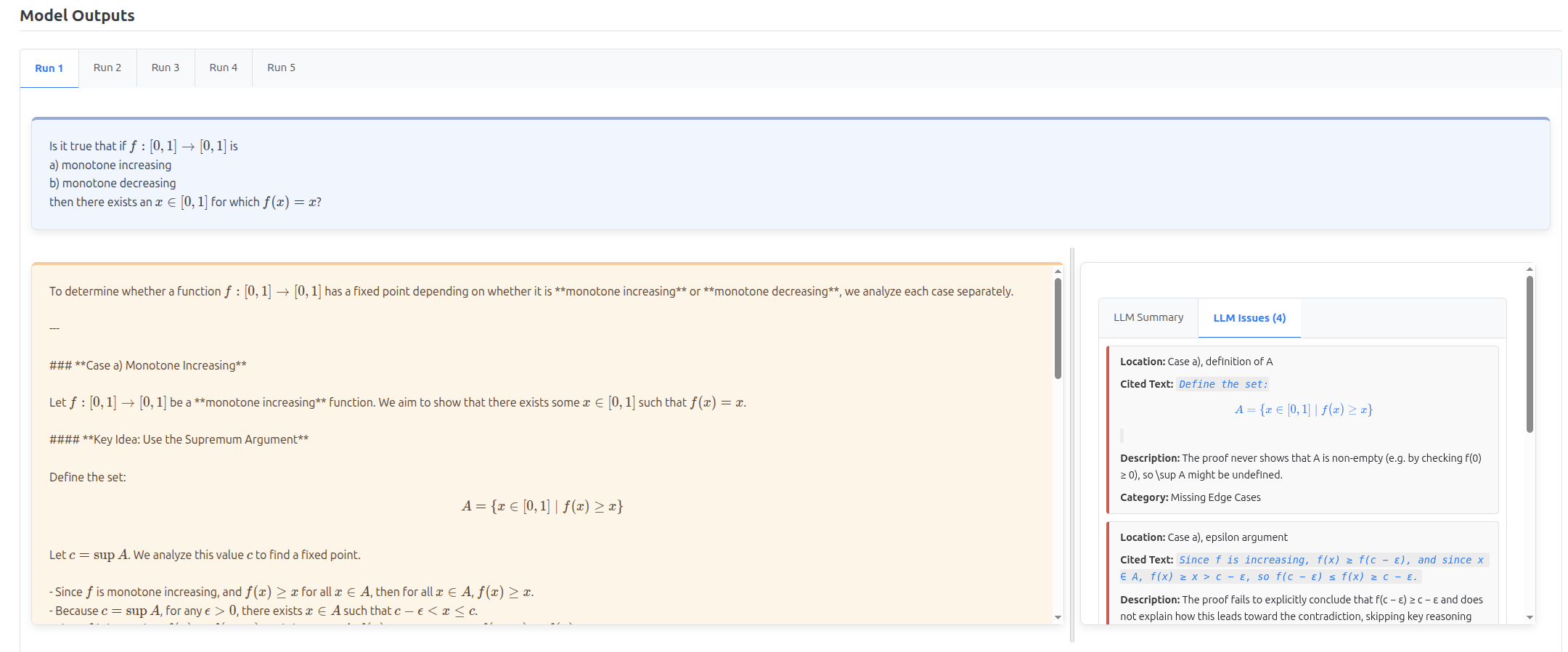}
    \caption{Interface for evaluating a generated solution. The problem is repeated for reference. The generated solution appears on the left, and the LLM's summary and identified issues on the right.}
    \label{fig:grading_interface_2}
\end{figure}

\begin{figure}[t]
    \centering
    \includegraphics[width=0.8\textwidth]{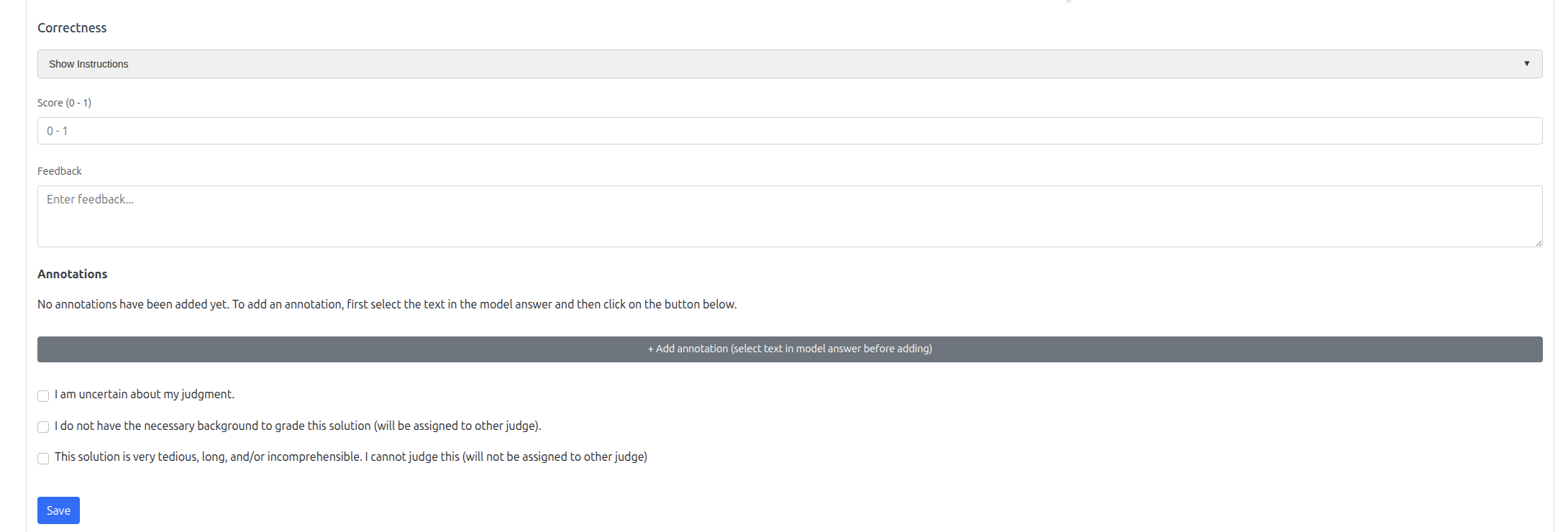}
    \caption{Grading form. Judges indicate correctness, provide a justification, highlight relevant content, and optionally express uncertainty or abstain.}
    \label{fig:grading_interface_3}
\end{figure}

\paragraph{Grading interface}  The grading interface was designed for clarity and ease of use. \cref{fig:grading_interface_1,fig:grading_interface_2,fig:grading_interface_3} illustrate its main components. The left panel contains a navigation bar for switching between problems and competitions. The right panel displays the problem statement and the ground-truth solution, along with options for flagging issues in either. Below, the generated solution is shown, accompanied by an automated summary and potential issues identified by an LLM judge. Judges can then evaluate the solution using a grading form that allows them to:
\vspace{-3mm}
\begin{itemize}\setlength\itemsep{0.005em}
    \item Indicate whether the solution is correct or incorrect

    \item Provide a brief justification
    
    \item Highlight specific parts of the solution relevant to their decision
    
    \item Indicate uncertainty or abstain from grading
\end{itemize}
\paragraph{Instructions}  Judges received a set of guidelines detailing how to use the interface and evaluate the correctness of solutions. Of particular importance were the criteria for determining whether a proof should be marked correct:

\begin{prompt}{Instructions for judges on when a proof is correct}
A solution should be considered correct even if it would earn 5+/7 points in a full grading. Examples of small penalties worth 1 point are if the solution: 
- Makes a small computational mistake that can be easily fixed
- Misses an edge case which can be easily proven/disproven
- Skips over a step that follows without much reasoning or manual work

A solution should be marked as incorrect if:
- It marks a step as trivial, if it is not immediately obvious why this would be the case
- It omits algebra-heavy computational steps, regardless of whether or not it has outlined the methodology
- Generalizes over a pattern without rigorously describing the pattern, or without proving any relevant properties.
- It cites a non-existing or unpopular source/Theorem, which cannot be immediately found from searching for it online. Thus, any theorems that can be immediately found and have a Wikipedia article are allowed.

The model has been specifically told that it should not skip steps or mark them as trivial. Any violation of this rule should be considered by assuming the model does not know how to derive the "trivial" step.
\end{prompt}
These instructions were developed collaboratively with the judges and refined iteratively based on their feedback, ensuring consistent grading across different problems and evaluators.

%% file: paper_files/app_prompts.tex
\clearpage
\section{Prompts}\label{app:prompts}

In this section, we provide the prompts used for various tasks in the OPC. The prompts are designed to be clear and concise, guiding the LLMs through the proof generation process while ensuring that they understand the requirements for correctness and clarity. In \cref{app:proof_generation_prompt}, we present the prompts used for generating proofs. In \cref{app:issues_interface_prompt}, we provide the prompt used to generate the LLM summary to aid human graders in identifying potential issues in the proof. In \cref{app:llm_as_judge_prompt}, we present the prompt used for LLMs to judge the correctness of a proof, used in \cref{sec:llmjudge}. In \cref{app:llm_as_discrete_judge_prompt,app:llm_as_continuous_judge_prompt,app:llm_as_rank_judge_prompt}, we provide all prompts used for the LLMs in best-of-n sampling, as described in \cref{sec:bestofn}.

\subsection{Proof Generation Prompt}\label{app:proof_generation_prompt}

The following prompt is used for problems with no final answer:
\begin{prompt}{Prompt}
Your task is to write a proof solution to the following problem. Your proof will be graded by human judges for accuracy, thoroughness, and clarity. When you write your proof, follow these guidelines:

- You are creating a proof, not a proof outline. Each step should be carefully explained and documented. If not properly explained, the judge will assume that you cannot explain it, and therefore decrease your grade.
- You can use general theorems and lemmas, but only if they are well-known. As a rule of thumb: if the result has a name and is famous enough to have a Wikipedia page or something similar to describe it, it is allowed. Any result from papers that would not be taught in high school or low-level bachelor courses in mathematics should not be used. Any use of such results will immediately give you a zero grade.
- Do not skip computation steps in your proof. Clearly explain what transformations were done and why they are allowed in each step of a calculation.
- You should use correct LaTeX notation to write equations and mathematical symbols. You should encompass these equations in appropriate symbols ("\\(" and "\\)" for inline math, "\\[" and "\\]" for block math) to enhance the clarity of your proof. Do not use any unicode characters.
- Your proof should be self-contained.
- If you are not sure about a specific step, or do not know how to prove an intermediate result, clearly state this. It is much preferable to indicate your uncertainty rather than making incorrect statements or claims.

{problem}
\end{prompt}

The following prompt is used for problems with a final answer:
\begin{prompt}{Prompt}
Your task is to write a proof solution to the following problem. Your proof will be graded by human judges for accuracy, thoroughness, and clarity. When you write your proof, follow these guidelines:

- You are creating a proof, not a proof outline. Each step should be carefully explained and documented. If not properly explained, the judge will assume that you cannot explain it, and therefore decrease your grade.
- You can use general theorems and lemmas, but only if they are well-known. As a rule of thumb: if the result has a name and is famous enough to have a Wikipedia page or something similar to describe it, it is allowed. Any result from papers that would not be taught in high school or low-level bachelor courses in mathematics should not be used. Any use of such results will immediately give you a zero grade.
- Do not skip computation steps in your proof. Clearly explain what transformations were done and why they are allowed in each step of a calculation.
- You should use correct LaTeX notation to write equations and mathematical symbols. You should encompass these equations in appropriate symbols ("\\(" and "\\)" for inline math, "\\[" and "\\]" for block math) to enhance the clarity of your proof. Do not use any unicode characters.
- Your proof should be self-contained.
- If you are not sure about a specific step, or do not know how to prove an intermediate result, clearly state this. It is much preferable to indicate your uncertainty rather than making incorrect statements or claims.
- Put your final answer within \\boxed{{}}.

{problem}
\end{prompt}

\subsection{Issues Interface Prompt}\label{app:issues_interface_prompt}

\begin{prompt}{Prompt}
Your task is to help a human mathematician grade a proof solution to the given problem. In this task, you will write a summary of the provided proof and highlight potential issues with it. 

### Input:

Your input will consist of the following components:
- **Problem Statement**: A mathematical problem that the proof is attempting to solve.
- **Ground-Truth Solution**: If available, the correct solution to the problem, which can be used as a reference. Note that ground-truth solutions may not always be provided, can also contain mistakes, and are often overly succinct. The ground-truth proof is mainly provided to help you understand the problem better.
- **Proof Solution**: The proof that you need to evaluate. This proof may contain errors, omissions, or unclear steps. The proof was generated by another language model, which was given the following instructions:
<model_prompt>
- You are creating a proof, not a proof outline. Each step should be carefully explained and documented. If not properly explained, the judge will assume that you cannot explain it, and therefore decrease your grade.
- You can use general theorems and lemmas, but only if they are well-known. As a rule of thumb: if the result has a name and is famous enough to have a Wikipedia page or something similar to describe it, it is allowed. Any result from papers that would not be taught in high school or low-level bachelor courses in mathematics should not be used. Any use of such results will immediately give you a zero grade.
- Do not skip computation steps in your proof. Clearly explain what transformations were done and why they are allowed in each step of a calculation.
- You should use correct LaTeX notation to write equations and mathematical symbols. You should encompass these equations in appropriate symbols ("\\(" and "\\)" for inline math, "\\[" and "\\]" for block math) to enhance the clarity of your proof. Do not use any unicode characters.
- Your proof should be self-contained.
- If you are not sure about a specific step, or do not know how to prove an intermediate result, clearly state this. It is much preferable to indicate your uncertainty rather than making incorrect statements or claims.
</model_prompt>

### Summary Guidelines:

First, you should write a concise summary of the proof solution. The summary should capture the main ideas and steps of the proof, but it does not need to be exhaustive. The goal is to provide a clear overview of what the proof is attempting to accomplish.
A summary should consist of only a few sentences, and it should not contain any judgment or evaluation of the proof. It should be purely descriptive.

### Potential Issues to Highlight:

Your main task is to identify potential issues in the proof solution. You should include any and all issues that you can find, no matter how small. Here are some common types of issues to look for:
- **Overgeneralization**: The generated proof proceeds by proving the problem in one or more specific cases, and then concludes that the result holds in general. However, it does not provide a proof for the general case.
- **Oversimplification**: The proof marks steps as trivial or obvious without proper justification. Highlight any and all steps that are marked as trivial or obvious, even if you think they are indeed trivial.
- **Skipping Computation Steps**: Proofs that skip computation steps or do not explain transformations clearly can lead to misunderstandings. Highlight any steps where the proof skips necessary calculations or explanations.
- **Citing Non-Standard Works or Theorems**: Some models may cite theorems or results that are not well-known or are not typically taught in high school or low-level bachelor courses. Such theorems are only allowed if they are well known. If the proof cites such results, highlight this as a potential issue, even if you think it is justified.
- **Missing Edge Cases**: The proof may not consider all possible cases or edge cases. If you notice that the proof does not address certain scenarios, highlight this as a potential issue.
- **Wrong Final Answer**: If the proof arrives at a final answer that is incorrect, highlight this as a potential issue.
- **Other**: Any other issues that do not fit into the above categories but you believe are significant enough to be highlighted.

For each of these issues, you should identify where in the proof they occur, provide a brief explanation of the issue, and indicate the category of the issue.

If there are more than four issues, you should only highlight the four most significant ones. Sort the issues by their significance, with the most significant issue first.

### Additional Instructions:

- Do not provide a final grade or score for the proof. Your task is to summarize and highlight potential issues, not to evaluate the proof as a whole.
- Be critical and thorough in your analysis. If you find no issues, you probably did not look closely enough.
- If you are unsure whether something is an issue, it is better to highlight it and let the human grader decide.
- Use clear and concise language in your summary and issue descriptions. The goal of your response is to help and speed up the human grader's work, not to add extra work for them. The more clear and concise your response is, the better it will be for the human grader.
- You should use correct LaTeX notation to write equations and mathematical symbols in your output JSON. You should encompass these equations in appropriate symbols ("\\(" and "\\)" for inline math, "\\[" and "\\]" for block math) to enhance the clarity of your proof. Do not use any unicode characters.
- Properly escape all symbols in your output JSON. For example, use `\\` for a single backslash.
- Spend special attention to producing valid JSON. It needs to be parsable by a standard JSON parser.

### Output Format:

Format your reply using a JSON object as follows:

```json
{{
"summary": "A concise summary of the proof solution.",
"issues": [
    {{
    "location": "A description of where the issue occurs in the proof",
    "text": "A citation or excerpt from the proof that contains the issue. If the issue is not contained to a very small part of the proof (e.g., a single sentence), you can leave this field empty.",
    "description": "A brief explanation of the issue.",
    "category": "The category of the issue (Overgeneralization, Oversimplification, Skipping Computation Steps, Citing Non-Standard Works or Theorems, Missing Edge Cases, Wrong Final Answer, Other)."
    }},
    ...
]
}}
```
If you truly cannot find any issues, you can return an empty issues array (either null or an empty list).

### Problem Statement:
{problem}

### Ground-Truth Solution:
{ground_truth_solution}

### Proof Solution:
{proof_solution}
\end{prompt}
\subsection{LLM as Judge Prompt}\label{app:llm_as_judge_prompt}
\begin{prompt}{Prompt}
You are judging the correctness of an LLM-generated proof for a math problem.

### Input:

Your input will consist of the following components:
- **Problem Statement**: A mathematical problem that the proof is attempting to solve.
- **Proof Solution**: The proof that you need to evaluate. This proof may contain errors, omissions, or unclear steps. The proof was generated by another language model, which was given the following instructions:
<model_prompt>
- You are creating a proof, not a proof outline. Each step should be carefully explained and documented. If not properly explained, the judge will assume that you cannot explain it, and therefore decrease your grade.
- You can use general theorems and lemmas, but only if they are well-known. As a rule of thumb: if the result has a name and is famous enough to have a Wikipedia page or something similar to describe it, it is allowed. Any result from papers that would not be taught in high school or low-level bachelor courses in mathematics should not be used. Any use of such results will immediately give you a zero grade.
- Do not skip computation steps in your proof. Clearly explain what transformations were done and why they are allowed in each step of a calculation.
- You should use correct LaTeX notation to write equations and mathematical symbols. You should encompass these equations in appropriate symbols ("\\(" and "\\)" for inline math, "\\[" and "\\]" for block math) to enhance the clarity of your proof. Do not use any unicode characters.
- Your proof should be self-contained.
- If you are not sure about a specific step, or do not know how to prove an intermediate result, clearly state this. It is much preferable to indicate your uncertainty rather than making incorrect statements or claims.
</model_prompt>

### How the solution should be graded:
A solution should be considered correct even if it would earn 5+/7 points in a standard grading format. Examples of small penalties worth 1 point are if the solution:
- Makes a small computational mistake that can be easily fixed
- Misses an edge case which can be easily proven/disproven
- Skips over a step that follows without much reasoning or manual work
Depending on the severity and the context, you may also not penalise a given error. On the other hand, a solution should be marked as incorrect if:
- It marks a step as trivial, if it is not immediately obvious with little reasoning why this would be the case.
- It omits algebra-heavy computational steps, regardless of whether or not it has outlined the methodology. Skipping shorter computations should be permitted.
- Generalizes over a pattern without rigorously describing the pattern, or without proving any relevant properties.
- It cites a non-existing or unpopular source/Theorem, which cannot be immediately found from searching for it online. Thus, any theorems that can be immediately found and have a Wikipedia article are allowed.

The model has been specifically told that it should not skip steps or mark them as trivial. Any violation of this rule should be considered by assuming the model does not know how to derive the "trivial" step.

### Scoring instructions

If you believe the proof is correct, end your analysis with \\boxed{{correct}}. If you believe the proof is incorrect, end your analysis with \\boxed{{incorrect}}.

### Problem Statement:
{problem}

### Model Solution:
{solution}
\end{prompt}

\subsection{LLM as Judge Prompt with Ground Truth Solution}\label{app:llm_as_judge_prompt_gt}
\begin{prompt}{Prompt}
You are judging the correctness of an LLM-generated proof for a math problem.

### Input:

Your input will consist of the following components:
- **Problem Statement**: A mathematical problem that the proof is attempting to solve.
- **Ground Truth Solution**: The solution of the problem, as originally written by the problem's authors.
- **Proof Solution**: The proof that you need to evaluate. This proof may contain errors, omissions, or unclear steps. The proof was generated by another language model, which was given the following instructions:
<model_prompt>
- You are creating a proof, not a proof outline. Each step should be carefully explained and documented. If not properly explained, the judge will assume that you cannot explain it, and therefore decrease your grade.
- You can use general theorems and lemmas, but only if they are well-known. As a rule of thumb: if the result has a name and is famous enough to have a Wikipedia page or something similar to describe it, it is allowed. Any result from papers that would not be taught in high-school or low-level bachelor courses in mathematics should not be used. Any use of such results will immediately give you a zero grade.
- Do not skip computation steps in your proof. Clearly explain what transformations were done and why they are allowed in each step of a calculation.
- You should use correct LaTeX notation to write equations and mathematical symbols. You should encompass these equations in appropriate symbols ("\\(" and "\\)" for inline math, "\\[" and "\\]" for block math) to enhance the clarity of your proof. Do not use any unicode characters.
- Your proof should be self-contained.
- If you are not sure about a specific step, or do not know how to prove an intermediate result, clearly state this. It is much preferable to indicate your uncertainty rather than making incorrect statements or claims.
</model_prompt>

### How the solution should be graded:
A solution should be considered correct even if it would earn 5+/7 points in a standard grading format. Examples of small penalties worth 1 point are if the solution:
- Makes a small computational mistake that can be easily fixed
- Misses an edge case which can be easily proven/disproven
- Skips over a step that follows without much reasoning or manual work
Depending on the severity and the context, you may also not penalise a given error. On the other hand, a solution should be marked as incorrect if:
- It marks a step as trivial, if it is not immediately obvious with little reasoning why this would be the case.
- It omits algebra-heavy computational steps, regardless of whether or not it has outlined the methodology. Skipping shorter computations should be permitted.
- Generalizes over a pattern without rigorously describing the pattern, or without proving any relevant properties.
- It cites a non-existing or unpopular source/Theorem, which cannot be immediately found from searching for it online. Thus, any theorems that can be immediately found and have a Wikipedia article are allowed.

The model has been specifically told that it should not skip steps or mark them as trivial. Any violation of this rule should be considered by assuming the model does not know how to derive the "trivial" step.

### Scoring instructions

If you believe the proof is correct, end your analysis with \\boxed{{correct}}. If you believe the proof is incorrect, end your analysis with \\boxed{{incorrect}}.

### Problem Statement:
{problem}

### Ground Truth Solution:
{gt_solution}

### Model Solution:
{solution}
\end{prompt}

\subsection{LLM as Discrete Judge Prompt}\label{app:llm_as_discrete_judge_prompt}
\begin{prompt}{Prompt}
You are judging the correctness of an LLM-generated proof for a math problem.

### Input:

Your input will consist of the following components:
- **Problem Statement**: A mathematical problem that the proof is attempting to solve.
- **Proof Solution**: The proof that you need to evaluate. This proof may contain errors, omissions, or unclear steps. The proof was generated by another language model, which was given the following instructions:
<model_prompt>
- You are creating a proof, not a proof outline. Each step should be carefully explained and documented. If not properly explained, the judge will assume that you cannot explain it, and therefore decrease your grade.
- You can use general theorems and lemmas, but only if they are well-known. As a rule of thumb: if the result has a name and is famous enough to have a Wikipedia page or something similar to describe it, it is allowed. Any result from papers that would not be taught in high school or low-level bachelor courses in mathematics should not be used. Any use of such results will immediately give you a zero grade.
- Do not skip computation steps in your proof. Clearly explain what transformations were done and why they are allowed in each step of a calculation.
- You should use correct LaTeX notation to write equations and mathematical symbols. You should encompass these equations in appropriate symbols ("\\(" and "\\)" for inline math, "\\[" and "\\]" for block math) to enhance the clarity of your proof. Do not use any unicode characters.
- Your proof should be self-contained.
- If you are not sure about a specific step, or do not know how to prove an intermediate result, clearly state this. It is much preferable to indicate your uncertainty rather than making incorrect statements or claims.
</model_prompt>

### How the solution should be graded:
A solution should be considered correct even if it would earn 5+/7 points in a standard grading format. Examples of small penalties worth 1 point are if the solution:
- Makes a small computational mistake that can be easily fixed
- Misses an edge case which can be easily proven/disproven
- Skips over a step that follows without much reasoning or manual work
Depending on the severity and the context, you may also not penalise a given error. On the other hand, a solution should be marked as incorrect if:
- It marks a step as trivial, if it is not immediately obvious with little reasoning why this would be the case.
- It omits algebra-heavy computational steps, regardless of whether or not it has outlined the methodology. Skipping shorter computations should be permitted.
- Generalizes over a pattern without rigorously describing the pattern, or without proving any relevant properties.
- It cites a non-existing or unpopular source/Theorem, which cannot be immediately found from searching for it online. Thus, any theorems that can be immediately found and have a Wikipedia article are allowed.

### Further Potential Issues:

Here are some common types of issues to look for:
- **Overgeneralization**: The generated proof proceeds by proving the problem in one or more specific cases, and then concludes that the result holds in general. However, it does not provide a proof for the general case.
- **Oversimplification**: The proof marks steps as trivial or obvious without proper justification.
- **Skipping Computation Steps**: Proofs that skip computation steps or do not explain transformations clearly can lead to misunderstandings.
- **Citing Non-Standard Works or Theorems**: Some models may cite theorems or results that are not well-known or are not typically taught in high school or low-level bachelor courses. Such theorems are only allowed if they are well known.
- **Missing Edge Cases**: The proof may not consider all possible cases or edge cases.

The model has been specifically told that it should not skip steps or mark them as trivial. Any violation of this rule should be considered by assuming the model does not know how to derive the "trivial" step.

### Scoring instructions

If you believe the proof is correct, end your analysis with \\boxed{{correct}}. If you believe the proof is incorrect, end your analysis with \\boxed{{incorrect}}.

### Problem Statement:
{problem}

### Model Solution:
{solution}
\end{prompt}

\subsection{LLM as Continuous Judge Prompt}\label{app:llm_as_continuous_judge_prompt}
\begin{prompt}{Prompt}
You are judging the correctness of an LLM-generated proof for a math problem.

### Input:

Your input will consist of the following components:
- **Problem Statement**: A mathematical problem that the proof is attempting to solve.
- **Proof Solution**: The proof that you need to evaluate. This proof may contain errors, omissions, or unclear steps. The proof was generated by another language model, which was given the following instructions:
<model_prompt>
- You are creating a proof, not a proof outline. Each step should be carefully explained and documented. If not properly explained, the judge will assume that you cannot explain it, and therefore decrease your grade.
- You can use general theorems and lemmas, but only if they are well-known. As a rule of thumb: if the result has a name and is famous enough to have a Wikipedia page or something similar to describe it, it is allowed. Any result from papers that would not be taught in high school or low-level bachelor courses in mathematics should not be used. Any use of such results will immediately give you a zero grade.
- Do not skip computation steps in your proof. Clearly explain what transformations were done and why they are allowed in each step of a calculation.
- You should use correct LaTeX notation to write equations and mathematical symbols. You should encompass these equations in appropriate symbols ("\\(" and "\\)" for inline math, "\\[" and "\\]" for block math) to enhance the clarity of your proof. Do not use any unicode characters.
- Your proof should be self-contained.
- If you are not sure about a specific step, or do not know how to prove an intermediate result, clearly state this. It is much preferable to indicate your uncertainty rather than making incorrect statements or claims.
</model_prompt>

### How the solution should be graded:
A solution should be graded out of a total of 7 points. Examples of small penalties worth 1 point are if the solution:
- Makes a small computational mistake that can be easily fixed
- Misses an edge case which can be easily proven/disproven
- Skips over a step that follows without much reasoning or manual work
Depending on the severity and the context, you may also not penalise a given error. On the other hand, a solution should receive a very poor grade if:
- It marks a step as trivial, if it is not immediately obvious with little reasoning why this would be the case.
- It omits algebra-heavy computational steps, regardless of whether or not it has outlined the methodology. Skipping shorter computations should be permitted.
- Generalizes over a pattern without rigorously describing the pattern, or without proving any relevant properties.
- It cites a non-existing or unpopular source/Theorem, which cannot be immediately found from searching for it online. Thus, any theorems that can be immediately found and have a Wikipedia article are allowed.

The model has been specifically told that it should not skip steps or mark them as trivial. Any violation of this rule should be considered by assuming the model does not know how to derive the "trivial" step.

### Further Potential Issues:

Here are some common types of issues to look for:
- **Overgeneralization**: The generated proof proceeds by proving the problem in one or more specific cases, and then concludes that the result holds in general. However, it does not provide a proof for the general case.
- **Oversimplification**: The proof marks steps as trivial or obvious without proper justification.
- **Skipping Computation Steps**: Proofs that skip computation steps or do not explain transformations clearly can lead to misunderstandings.
- **Citing Non-Standard Works or Theorems**: Some models may cite theorems or results that are not well-known or are not typically taught in high school or low-level bachelor courses. Such theorems are only allowed if they are well known.
- **Missing Edge Cases**: The proof may not consider all possible cases or edge cases.

### Scoring instructions

Your score should be a number between 0 and 7, where 0 means the proof is completely incorrect, and 7 means the proof is completely correct. Be very critical in your grading. If you find small errors, deduct points accordingly.

### Output Format:

At the end of your analysis, present your grade as a number between 0 and 7 in "$\boxed{{}}$".

### Problem Statement:
{problem}

### Model Solution:
{solution}
\end{prompt}

\subsection{LLM as Rank Judge Prompt}\label{app:llm_as_rank_judge_prompt}

\begin{prompt}{Prompt}
You are judging which of the two LLM-generated proofs for a given math problem is better.

### Input:

Your input will consist of the following components:
- **Problem Statement**: A mathematical problem that the proof is attempting to solve.
- **Proof Solution A/B**: The proofs that you need to evaluate. This proof may contain errors, omissions, or unclear steps. Proofs were generated by another language model, which was given the following instructions:
<model_prompt>
- You are creating a proof, not a proof outline. Each step should be carefully explained and documented. If not properly explained, the judge will assume that you cannot explain it, and therefore decrease your grade.
- You can use general theorems and lemmas, but only if they are well-known. As a rule of thumb: if the result has a name and is famous enough to have a Wikipedia page or something similar to describe it, it is allowed. Any result from papers that would not be taught in high school or low-level bachelor courses in mathematics should not be used. Any use of such results will immediately give you a zero grade.
- Do not skip computation steps in your proof. Clearly explain what transformations were done and why they are allowed in each step of a calculation.
- You should use correct LaTeX notation to write equations and mathematical symbols. You should encompass these equations in appropriate symbols ("\\(" and "\\)" for inline math, "\\[" and "\\]" for block math) to enhance the clarity of your proof. Do not use any unicode characters.
- Your proof should be self-contained.
- If you are not sure about a specific step, or do not know how to prove an intermediate result, clearly state this. It is much preferable to indicate your uncertainty rather than making incorrect statements or claims.
</model_prompt>

### How the solution should be graded:
The following examples are small mistakes that should only be slightly penalised:
- Makes a small computational mistake that can be easily fixed
- Misses an edge case which can be easily proven/disproven
- Skips over a step that follows without much reasoning or manual work
On the other hand, a solution should should be severely penalised if:
- It marks a step as trivial, if it is not immediately obvious with little reasoning why this would be the case.
- It omits algebra-heavy computational steps, regardless of whether or not it has outlined the methodology. Skipping shorter computations should be permitted.
- Generalizes over a pattern without rigorously describing the pattern, or without proving any relevant properties.
- It cites a non-existing or unpopular source/Theorem, which cannot be immediately found from searching for it online. Thus, any theorems that can be immediately found and have a Wikipedia article are allowed.

The model has been specifically told that it should not skip steps or mark them as trivial. Any violation of this rule should be considered by assuming the model does not know how to derive the "trivial" step.

### Further Potential Issues:

Here are some common types of issues to look for:
- **Overgeneralization**: The generated proof proceeds by proving the problem in one or more specific cases, and then concludes that the result holds in general. However, it does not provide a proof for the general case.
- **Oversimplification**: The proof marks steps as trivial or obvious without proper justification.
- **Skipping Computation Steps**: Proofs that skip computation steps or do not explain transformations clearly can lead to misunderstandings.
- **Citing Non-Standard Works or Theorems**: Some models may cite theorems or results that are not well-known or are not typically taught in high school or low-level bachelor courses. Such theorems are only allowed if they are well known.
- **Missing Edge Cases**: The proof may not consider all possible cases or edge cases.

### Scoring instructions

You should compare the two proofs and determine which one is better. If you believe Proof A is better, end your analysis with \\boxed{{A}}. If you believe Proof B is better, end your analysis with \\boxed{{B}}. If you believe both proofs are equally good, end your analysis with \\boxed{{equal}}.

### Problem Statement:
{problem}

### Proof Solution A:
{solution_a}

### Proof Solution B:
{solution_b}
\end{prompt}

%% file: references.bib
@article{r1,
  title={Deepseek-r1: Incentivizing reasoning capability in llms via reinforcement learning},
  author={Guo, Daya and Yang, Dejian and Zhang, Haowei and Song, Junxiao and Zhang, Ruoyu and Xu, Runxin and Zhu, Qihao and Ma, Shirong and Wang, Peiyi and Bi, Xiao and others},
  journal={arXiv preprint arXiv:2501.12948},
  year={2025}
}

@article{grpo,
  author       = {Zhihong Shao and
                  Peiyi Wang and
                  Qihao Zhu and
                  Runxin Xu and
                  Junxiao Song and
                  Mingchuan Zhang and
                  Y. K. Li and
                  Y. Wu and
                  Daya Guo},
  title        = {DeepSeekMath: Pushing the Limits of Mathematical Reasoning in Open
                  Language Models},
  journal      = {CoRR},
  volume       = {abs/2402.03300},
  year         = {2024},
  url          = {https://doi.org/10.48550/arXiv.2402.03300},
  doi          = {10.48550/ARXIV.2402.03300},
  eprinttype    = {arXiv},
  eprint       = {2402.03300},
  bibsource    = {dblp computer science bibliography, https://dblp.org}
}

@misc{matharena,
	title = {MathArena: Evaluating LLMs on Uncontaminated Math Competitions},
  author = {Mislav Balunović and Jasper Dekoninck and Ivo Petrov and Nikola Jovanović and Martin Vechev},
	copyright = {MIT},
	url = {https://matharena.ai/},
	publisher = {SRI Lab, ETH Zurich},
	month = feb,
	year = {2025},
}

@inproceedings{putnambench,
  author       = {George Tsoukalas and
                  Jasper Lee and
                  John Jennings and
                  Jimmy Xin and
                  Michelle Ding and
                  Michael Jennings and
                  Amitayush Thakur and
                  Swarat Chaudhuri},
  editor       = {Amir Globersons and
                  Lester Mackey and
                  Danielle Belgrave and
                  Angela Fan and
                  Ulrich Paquet and
                  Jakub M. Tomczak and
                  Cheng Zhang},
  title        = {PutnamBench: Evaluating Neural Theorem-Provers on the Putnam Mathematical
                  Competition},
  booktitle    = {Advances in Neural Information Processing Systems 38: Annual Conference
                  on Neural Information Processing Systems 2024, NeurIPS 2024, Vancouver,
                  BC, Canada, December 10 - 15, 2024},
  year         = {2024},
  url          = {http://papers.nips.cc/paper\_files/paper/2024/hash/1582eaf9e0cf349e1e5a6ee453100aa1-Abstract-Datasets\_and\_Benchmarks\_Track.html},
  bibsource    = {dblp computer science bibliography, https://dblp.org}
}

@inproceedings{lean,
  author       = {Leonardo de Moura and
                  Sebastian Ullrich},
  title        = {The Lean 4 Theorem Prover and Programming Language},
  booktitle    = {{CADE}},
  series       = {Lecture Notes in Computer Science},
  volume       = {12699},
  pages        = {625--635},
  publisher    = {Springer},
  year         = {2021}
}

@inproceedings{minif2f,
  author       = {Kunhao Zheng and
                  Jesse Michael Han and
                  Stanislas Polu},
  title        = {miniF2F: a cross-system benchmark for formal Olympiad-level mathematics},
  booktitle    = {The Tenth International Conference on Learning Representations, {ICLR}
                  2022, Virtual Event, April 25-29, 2022},
  publisher    = {OpenReview.net},
  year         = {2022},
  url          = {https://openreview.net/forum?id=9ZPegFuFTFv},
  bibsource    = {dblp computer science bibliography, https://dblp.org}
}

@misc{o3,
  title={OpenAI o3-mini System Card},
  author={OpenAI},
  pages={37},
  year={2025},
  month = {January},
  url = {https://cdn.openai.com/o3-mini-system-card-feb10.pdf},

}

@misc{xai2025grok3,
  author       = {{xAI}},
  title        = {Grok 3 Beta — The Age of Reasoning Agents},
  year         = {2025},
  month        = feb,
  url          = {https://x.ai/news/grok-3},
  note         = {Accessed: 2025-04-03}
}

@article{o1,
  author       = {Aaron Jaech and
                  Adam Kalai and
                  Adam Lerer and
                  Adam Richardson and
                  Ahmed El{-}Kishky and
                  Aiden Low and
                  Alec Helyar and
                  Aleksander Madry and
                  Alex Beutel and
                  Alex Carney and
                  Alex Iftimie and
                  Alex Karpenko and
                  Alex Tachard Passos and
                  Alexander Neitz and
                  Alexander Prokofiev and
                  Alexander Wei and
                  Allison Tam and
                  Ally Bennett and
                  Ananya Kumar and
                  Andre Saraiva and
                  Andrea Vallone and
                  Andrew Duberstein and
                  Andrew Kondrich and
                  Andrey Mishchenko and
                  Andy Applebaum and
                  Angela Jiang and
                  Ashvin Nair and
                  Barret Zoph and
                  Behrooz Ghorbani and
                  Ben Rossen and
                  Benjamin Sokolowsky and
                  Boaz Barak and
                  Bob McGrew and
                  Borys Minaiev and
                  Botao Hao and
                  Bowen Baker and
                  Brandon Houghton and
                  Brandon McKinzie and
                  Brydon Eastman and
                  Camillo Lugaresi and
                  Cary Bassin and
                  Cary Hudson and
                  Chak Ming Li and
                  Charles de Bourcy and
                  Chelsea Voss and
                  Chen Shen and
                  Chong Zhang and
                  Chris Koch and
                  Chris Orsinger and
                  Christopher Hesse and
                  Claudia Fischer and
                  Clive Chan and
                  Dan Roberts and
                  Daniel Kappler and
                  Daniel Levy and
                  Daniel Selsam and
                  David Dohan and
                  David Farhi and
                  David Mely and
                  David Robinson and
                  Dimitris Tsipras and
                  Doug Li and
                  Dragos Oprica and
                  Eben Freeman and
                  Eddie Zhang and
                  Edmund Wong and
                  Elizabeth Proehl and
                  Enoch Cheung and
                  Eric Mitchell and
                  Eric Wallace and
                  Erik Ritter and
                  Evan Mays and
                  Fan Wang and
                  Felipe Petroski Such and
                  Filippo Raso and
                  Florencia Leoni and
                  Foivos Tsimpourlas and
                  Francis Song and
                  Fred von Lohmann and
                  Freddie Sulit and
                  Geoff Salmon and
                  Giambattista Parascandolo and
                  Gildas Chabot and
                  Grace Zhao and
                  Greg Brockman and
                  Guillaume Leclerc and
                  Hadi Salman and
                  Haiming Bao and
                  Hao Sheng and
                  Hart Andrin and
                  Hessam Bagherinezhad and
                  Hongyu Ren and
                  Hunter Lightman and
                  Hyung Won Chung and
                  Ian Kivlichan and
                  Ian O'Connell and
                  Ian Osband and
                  Ignasi Clavera Gilaberte and
                  Ilge Akkaya},
  title        = {OpenAI o1 System Card},
  journal      = {CoRR},
  volume       = {abs/2412.16720},
  year         = {2024},
  url          = {https://doi.org/10.48550/arXiv.2412.16720},
  doi          = {10.48550/ARXIV.2412.16720},
  eprinttype    = {arXiv},
  eprint       = {2412.16720},
  bibsource    = {dblp computer science bibliography, https://dblp.org}
}

@article{deepmind2025geminipro,
  author       = {Gemini Team},
  title        = {Gemini 2.5: Pushing the Frontier with Advanced Reasoning, Multimodality,
                  Long Context, and Next Generation Agentic Capabilities},
  journal      = {CoRR},
  volume       = {abs/2507.06261},
  year         = {2025},
  url          = {https://doi.org/10.48550/arXiv.2507.06261},
  doi          = {10.48550/ARXIV.2507.06261},
  eprinttype    = {arXiv},
  eprint       = {2507.06261},
  bibsource    = {dblp computer science bibliography, https://dblp.org}
}

@inproceedings{frieder2024,
  author       = {Simon Frieder and
                  Luca Pinchetti and
                  Alexis Chevalier and
                  Ryan{-}Rhys Griffiths and
                  Tommaso Salvatori and
                  Thomas Lukasiewicz and
                  Philipp Petersen and
                  Julius Berner},
  editor       = {Alice Oh and
                  Tristan Naumann and
                  Amir Globerson and
                  Kate Saenko and
                  Moritz Hardt and
                  Sergey Levine},
  title        = {Mathematical Capabilities of ChatGPT},
  booktitle    = {Advances in Neural Information Processing Systems 36: Annual Conference
                  on Neural Information Processing Systems 2023, NeurIPS 2023, New Orleans,
                  LA, USA, December 10 - 16, 2023},
  year         = {2023},
  url          = {http://papers.nips.cc/paper\_files/paper/2023/hash/58168e8a92994655d6da3939e7cc0918-Abstract-Datasets\_and\_Benchmarks.html},
  bibsource    = {dblp computer science bibliography, https://dblp.org}
}

@article{brainvbytes,
  author       = {Hamed Mahdavi and
                  Alireza Hashemi and
                  Majid Daliri and
                  Pegah Mohammadipour and
                  Alireza Farhadi and
                  Samira Malek and
                  Yekta Yazdanifard and
                  Amir Khasahmadi and
                  Vasant G. Honavar},
  title        = {Brains vs. Bytes: Evaluating {LLM} Proficiency in Olympiad Mathematics},
  journal      = {CoRR},
  volume       = {abs/2504.01995},
  year         = {2025},
  url          = {https://doi.org/10.48550/arXiv.2504.01995},
  doi          = {10.48550/ARXIV.2504.01995},
  eprinttype    = {arXiv},
  eprint       = {2504.01995},
  bibsource    = {dblp computer science bibliography, https://dblp.org}
}

@article{gsm8k,
  author       = {Karl Cobbe and
                  Vineet Kosaraju and
                  Mohammad Bavarian and
                  Mark Chen and
                  Heewoo Jun and
                  Lukasz Kaiser and
                  Matthias Plappert and
                  Jerry Tworek and
                  Jacob Hilton and
                  Reiichiro Nakano and
                  Christopher Hesse and
                  John Schulman},
  title        = {Training Verifiers to Solve Math Word Problems},
  journal      = {CoRR},
  volume       = {abs/2110.14168},
  year         = {2021},
  url          = {https://arxiv.org/abs/2110.14168},
  eprinttype    = {arXiv},
  eprint       = {2110.14168},
  bibsource    = {dblp computer science bibliography, https://dblp.org}
}

@inproceedings{math500,
  author       = {Hunter Lightman and
                  Vineet Kosaraju and
                  Yuri Burda and
                  Harrison Edwards and
                  Bowen Baker and
                  Teddy Lee and
                  Jan Leike and
                  John Schulman and
                  Ilya Sutskever and
                  Karl Cobbe},
  title        = {Let's Verify Step by Step},
  booktitle    = {The Twelfth International Conference on Learning Representations,
                  {ICLR} 2024, Vienna, Austria, May 7-11, 2024},
  publisher    = {OpenReview.net},
  year         = {2024},
  url          = {https://openreview.net/forum?id=v8L0pN6EOi},
  bibsource    = {dblp computer science bibliography, https://dblp.org}
}

@inproceedings{omnimath,
  author       = {Bofei Gao and
                  Feifan Song and
                  Zhe Yang and
                  Zefan Cai and
                  Yibo Miao and
                  Qingxiu Dong and
                  Lei Li and
                  Chenghao Ma and
                  Liang Chen and
                  Runxin Xu and
                  Zhengyang Tang and
                  Benyou Wang and
                  Daoguang Zan and
                  Shanghaoran Quan and
                  Ge Zhang and
                  Lei Sha and
                  Yichang Zhang and
                  Xuancheng Ren and
                  Tianyu Liu and
                  Baobao Chang},
  title        = {Omni-MATH: {A} Universal Olympiad Level Mathematic Benchmark for Large
                  Language Models},
  booktitle    = {The Thirteenth International Conference on Learning Representations,
                  {ICLR} 2025, Singapore, April 24-28, 2025},
  publisher    = {OpenReview.net},
  year         = {2025},
  url          = {https://openreview.net/forum?id=yaqPf0KAlN},
  bibsource    = {dblp computer science bibliography, https://dblp.org}
}

@article{frontiermath,
  author       = {Elliot Glazer and
                  Ege Erdil and
                  Tamay Besiroglu and
                  Diego Chicharro and
                  Evan Chen and
                  Alex Gunning and
                  Caroline Falkman Olsson and
                  Jean{-}Stanislas Denain and
                  Anson Ho and
                  Emily de Oliveira Santos and
                  Olli J{\"{a}}rviniemi and
                  Matthew Barnett and
                  Robert Sandler and
                  Matej Vrzala and
                  Jaime Sevilla and
                  Qiuyu Ren and
                  Elizabeth Pratt and
                  Lionel Levine and
                  Grant Barkley and
                  Natalie Stewart and
                  Bogdan Grechuk and
                  Tetiana Grechuk and
                  Shreepranav Varma Enugandla and
                  Mark Wildon},
  title        = {FrontierMath: {A} Benchmark for Evaluating Advanced Mathematical Reasoning
                  in {AI}},
  journal      = {CoRR},
  volume       = {abs/2411.04872},
  year         = {2024},
  url          = {https://doi.org/10.48550/arXiv.2411.04872},
  doi          = {10.48550/ARXIV.2411.04872},
  eprinttype    = {arXiv},
  eprint       = {2411.04872},
  bibsource    = {dblp computer science bibliography, https://dblp.org}
}

@inproceedings{olympiadbench,
  author       = {Chaoqun He and
                  Renjie Luo and
                  Yuzhuo Bai and
                  Shengding Hu and
                  Zhen Leng Thai and
                  Junhao Shen and
                  Jinyi Hu and
                  Xu Han and
                  Yujie Huang and
                  Yuxiang Zhang and
                  Jie Liu and
                  Lei Qi and
                  Zhiyuan Liu and
                  Maosong Sun},
  editor       = {Lun{-}Wei Ku and
                  Andre Martins and
                  Vivek Srikumar},
  title        = {OlympiadBench: {A} Challenging Benchmark for Promoting {AGI} with
                  Olympiad-Level Bilingual Multimodal Scientific Problems},
  booktitle    = {Proceedings of the 62nd Annual Meeting of the Association for Computational
                  Linguistics (Volume 1: Long Papers), {ACL} 2024, Bangkok, Thailand,
                  August 11-16, 2024},
  pages        = {3828--3850},
  publisher    = {Association for Computational Linguistics},
  year         = {2024},
  url          = {https://doi.org/10.18653/v1/2024.acl-long.211},
  doi          = {10.18653/V1/2024.ACL-LONG.211},
  bibsource    = {dblp computer science bibliography, https://dblp.org}
}

@article{rstar,
  author       = {Xinyu Guan and
                  Li Lyna Zhang and
                  Yifei Liu and
                  Ning Shang and
                  Youran Sun and
                  Yi Zhu and
                  Fan Yang and
                  Mao Yang},
  title        = {rStar-Math: Small LLMs Can Master Math Reasoning with Self-Evolved
                  Deep Thinking},
  journal      = {CoRR},
  volume       = {abs/2501.04519},
  year         = {2025},
  url          = {https://doi.org/10.48550/arXiv.2501.04519},
  doi          = {10.48550/ARXIV.2501.04519},
  eprinttype    = {arXiv},
  eprint       = {2501.04519},
  bibsource    = {dblp computer science bibliography, https://dblp.org}
}

@misc{deeptheorem,
      title={DeepTheorem: Advancing LLM Reasoning for Theorem Proving Through Natural Language and Reinforcement Learning}, 
      author={Ziyin Zhang and Jiahao Xu and Zhiwei He and Tian Liang and Qiuzhi Liu and Yansi Li and Linfeng Song and Zhenwen Liang and Zhuosheng Zhang and Rui Wang and Zhaopeng Tu and Haitao Mi and Dong Yu},
      year={2025},
      eprint={2505.23754},
      archivePrefix={arXiv},
      primaryClass={cs.CL},
      url={https://arxiv.org/abs/2505.23754}, 
}

@article{deepseekproverv2,
  author       = {Z. Z. Ren and
                  Zhihong Shao and
                  Junxiao Song and
                  Huajian Xin and
                  Haocheng Wang and
                  Wanjia Zhao and
                  Liyue Zhang and
                  Zhe Fu and
                  Qihao Zhu and
                  Dejian Yang and
                  Z. F. Wu and
                  Zhibin Gou and
                  Shirong Ma and
                  Hongxuan Tang and
                  Yuxuan Liu and
                  Wenjun Gao and
                  Daya Guo and
                  Chong Ruan},
  title        = {DeepSeek-Prover-V2: Advancing Formal Mathematical Reasoning via Reinforcement
                  Learning for Subgoal Decomposition},
  journal      = {CoRR},
  volume       = {abs/2504.21801},
  year         = {2025},
  url          = {https://doi.org/10.48550/arXiv.2504.21801},
  doi          = {10.48550/ARXIV.2504.21801},
  eprinttype    = {arXiv},
  eprint       = {2504.21801},
  bibsource    = {dblp computer science bibliography, https://dblp.org}
}

@article{prooforbluff,
  author       = {Ivo Petrov and
                  Jasper Dekoninck and
                  Lyuben Baltadzhiev and
                  Maria Drencheva and
                  Kristian Minchev and
                  Mislav Balunovic and
                  Nikola Jovanovic and
                  Martin T. Vechev},
  title        = {Proof or Bluff? Evaluating LLMs on 2025 {USA} Math Olympiad},
  journal      = {CoRR},
  volume       = {abs/2503.21934},
  year         = {2025},
  url          = {https://doi.org/10.48550/arXiv.2503.21934},
  doi          = {10.48550/ARXIV.2503.21934},
  eprinttype    = {arXiv},
  eprint       = {2503.21934},
  bibsource    = {dblp computer science bibliography, https://dblp.org}
}

@article{kimina,
  author       = {Haiming Wang and
                  Mert Unsal and
                  Xiaohan Lin and
                  Mantas Baksys and
                  Junqi Liu and
                  Marco Dos Santos and
                  Flood Sung and
                  Marina Vinyes and
                  Zhenzhe Ying and
                  Zekai Zhu and
                  Jianqiao Lu and
                  Hugues de Saxc{\'{e}} and
                  Bolton Bailey and
                  Chendong Song and
                  Chenjun Xiao and
                  Dehao Zhang and
                  Ebony Zhang and
                  Frederick Pu and
                  Han Zhu and
                  Jiawei Liu and
                  Jonas Bayer and
                  Julien Michel and
                  Longhui Yu and
                  L{\'{e}}o Dreyfus{-}Schmidt and
                  Lewis Tunstall and
                  Luigi Pagani and
                  Moreira Machado and
                  Pauline Bourigault and
                  Ran Wang and
                  Stanislas Polu and
                  Thibaut Barroyer and
                  Wen{-}Ding Li and
                  Yazhe Niu and
                  Yann Fleureau and
                  Yangyang Hu and
                  Zhouliang Yu and
                  Zihan Wang and
                  Zhilin Yang and
                  Zhengying Liu and
                  Jia Li},
  title        = {Kimina-Prover Preview: Towards Large Formal Reasoning Models with
                  Reinforcement Learning},
  journal      = {CoRR},
  volume       = {abs/2504.11354},
  year         = {2025},
  url          = {https://doi.org/10.48550/arXiv.2504.11354},
  doi          = {10.48550/ARXIV.2504.11354},
  eprinttype    = {arXiv},
  eprint       = {2504.11354},
  bibsource    = {dblp computer science bibliography, https://dblp.org}
}

@misc{numina_math_datasets,
  author = {Jia Li and Edward Beeching and Lewis Tunstall and Ben Lipkin and Roman Soletskyi and Shengyi Costa Huang and Kashif Rasul and Longhui Yu and Albert Jiang and Ziju Shen and Zihan Qin and Bin Dong and Li Zhou and Yann Fleureau and Guillaume Lample and Stanislas Polu},
  title = {NuminaMath},
  year = {2024},
  publisher = {Numina},
  journal = {Hugging Face repository},
  howpublished = {\url{[https://huggingface.co/AI-MO/NuminaMath-1.5](https://github.com/project-numina/aimo-progress-prize/blob/main/report/numina_dataset.pdf)}}
}

@article{aimo2,
  author       = {Ivan Moshkov and
                  Darragh Hanley and
                  Ivan Sorokin and
                  Shubham Toshniwal and
                  Christof Henkel and
                  Benedikt Schifferer and
                  Wei Du and
                  Igor Gitman},
  title        = {{AIMO-2} Winning Solution: Building State-of-the-Art Mathematical
                  Reasoning Models with OpenMathReasoning dataset},
  journal      = {CoRR},
  volume       = {abs/2504.16891},
  year         = {2025},
  url          = {https://doi.org/10.48550/arXiv.2504.16891},
  doi          = {10.48550/ARXIV.2504.16891},
  eprinttype    = {arXiv},
  eprint       = {2504.16891},
  bibsource    = {dblp computer science bibliography, https://dblp.org}
}

@misc{bigmath,
      title={Big-Math: A Large-Scale, High-Quality Math Dataset for Reinforcement Learning in Language Models}, 
      author={Alon Albalak and Duy Phung and Nathan Lile and Rafael Rafailov and Kanishk Gandhi and Louis Castricato and Anikait Singh and Chase Blagden and Violet Xiang and Dakota Mahan and Nick Haber},
      year={2025},
      eprint={2502.17387},
      archivePrefix={arXiv},
      primaryClass={cs.LG},
      url={https://arxiv.org/abs/2502.17387}, 
}

@article{deepmath,
  author       = {Zhiwei He and
                  Tian Liang and
                  Jiahao Xu and
                  Qiuzhi Liu and
                  Xingyu Chen and
                  Yue Wang and
                  Linfeng Song and
                  Dian Yu and
                  Zhenwen Liang and
                  Wenxuan Wang and
                  Zhuosheng Zhang and
                  Rui Wang and
                  Zhaopeng Tu and
                  Haitao Mi and
                  Dong Yu},
  title        = {DeepMath-103K: {A} Large-Scale, Challenging, Decontaminated, and Verifiable
                  Mathematical Dataset for Advancing Reasoning},
  journal      = {CoRR},
  volume       = {abs/2504.11456},
  year         = {2025},
  url          = {https://doi.org/10.48550/arXiv.2504.11456},
  doi          = {10.48550/ARXIV.2504.11456},
  eprinttype    = {arXiv},
  eprint       = {2504.11456},
  bibsource    = {dblp computer science bibliography, https://dblp.org}
}

@misc{o4,
  title={OpenAI o3 and o4-mini System Card},
  author={OpenAI},
  pages={33},
  year={2025},
  month = {April},
  url = {https://cdn.openai.com/pdf/2221c875-02dc-4789-800b-e7758f3722c1/o3-and-o4-mini-system-card.pdf},
}

@misc{qwen3blog2025,
  author      = {{Qwen Team}},
  title       = {{Qwen3: Think Deeper, Act Faster}},
  year        = {2025},
  month       = apr,
  day         = {29},
  url         = {https://qwenlm.github.io/blog/qwen3/},
  urldate     = {2025-06-16},
  note        = {Qwen Blog}
}

@misc{knockout,
      title={PairJudge RM: Perform Best-of-N Sampling with Knockout Tournament}, 
      author={Yantao Liu and Zijun Yao and Rui Min and Yixin Cao and Lei Hou and Juanzi Li},
      year={2025},
      eprint={2501.13007},
      archivePrefix={arXiv},
      primaryClass={cs.CL},
      url={https://arxiv.org/abs/2501.13007}, 
}

@article{bradley1952rank,
    author = {Bradley, Ralph Allan and Terry, Milton E.},
    title={Rank Analysis of Incomplete Block Designs: The Method of Paired Comparisons},
    journal = {Biometrika},
    volume = {39},
    number = {3-4},
    pages = {324-345},
    year = {1952},
    month = {12},
    issn = {0006-3444},
    doi = {10.1093/biomet/39.3-4.324},
    url = {https://doi.org/10.1093/biomet/39.3-4.324},
    eprint = {https://academic.oup.com/biomet/article-pdf/39/3-4/324/930466/39-3-4-324.pdf},
}

@article{llmasjudgesurvey,
  author       = {Jiawei Gu and
                  Xuhui Jiang and
                  Zhichao Shi and
                  Hexiang Tan and
                  Xuehao Zhai and
                  Chengjin Xu and
                  Wei Li and
                  Yinghan Shen and
                  Shengjie Ma and
                  Honghao Liu and
                  Yuanzhuo Wang and
                  Jian Guo},
  title        = {A Survey on LLM-as-a-Judge},
  journal      = {CoRR},
  volume       = {abs/2411.15594},
  year         = {2024},
  url          = {https://doi.org/10.48550/arXiv.2411.15594},
  doi          = {10.48550/ARXIV.2411.15594},
  eprinttype    = {arXiv},
  eprint       = {2411.15594},
  bibsource    = {dblp computer science bibliography, https://dblp.org}
}

@inproceedings{llmasajudge,
  author       = {Lianmin Zheng and
                  Wei{-}Lin Chiang and
                  Ying Sheng and
                  Siyuan Zhuang and
                  Zhanghao Wu and
                  Yonghao Zhuang and
                  Zi Lin and
                  Zhuohan Li and
                  Dacheng Li and
                  Eric P. Xing and
                  Hao Zhang and
                  Joseph E. Gonzalez and
                  Ion Stoica},
  editor       = {Alice Oh and
                  Tristan Naumann and
                  Amir Globerson and
                  Kate Saenko and
                  Moritz Hardt and
                  Sergey Levine},
  title        = {Judging LLM-as-a-Judge with MT-Bench and Chatbot Arena},
  booktitle    = {Advances in Neural Information Processing Systems 36: Annual Conference
                  on Neural Information Processing Systems 2023, NeurIPS 2023, New Orleans,
                  LA, USA, December 10 - 16, 2023},
  year         = {2023},
  url          = {http://papers.nips.cc/paper\_files/paper/2023/hash/91f18a1287b398d378ef22505bf41832-Abstract-Datasets\_and\_Benchmarks.html},
  bibsource    = {dblp computer science bibliography, https://dblp.org}
}

@misc{ineqmath,
      title={Solving Inequality Proofs with Large Language Models}, 
      author={Jiayi Sheng and Luna Lyu and Jikai Jin and Tony Xia and Alex Gu and James Zou and Pan Lu},
      year={2025},
      eprint={2506.07927},
      archivePrefix={arXiv},
      primaryClass={cs.AI},
      url={https://arxiv.org/abs/2506.07927}, 
}

@misc{rightisnotenough,
      title={Right Is Not Enough: The Pitfalls of Outcome Supervision in Training LLMs for Math Reasoning}, 
      author={Jiaxing Guo and Wenjie Yang and Shengzhong Zhang and Tongshan Xu and Lun Du and Da Zheng and Zengfeng Huang},
      year={2025},
      eprint={2506.06877},
      archivePrefix={arXiv},
      primaryClass={cs.CL},
      url={https://arxiv.org/abs/2506.06877}, 
}

@article{goedelprover,
  author       = {Yong Lin and
                  Shange Tang and
                  Bohan Lyu and
                  Jiayun Wu and
                  Hongzhou Lin and
                  Kaiyu Yang and
                  Jia Li and
                  Mengzhou Xia and
                  Danqi Chen and
                  Sanjeev Arora and
                  Chi Jin},
  title        = {Goedel-Prover: {A} Frontier Model for Open-Source Automated Theorem
                  Proving},
  journal      = {CoRR},
  volume       = {abs/2502.07640},
  year         = {2025},
  url          = {https://doi.org/10.48550/arXiv.2502.07640},
  doi          = {10.48550/ARXIV.2502.07640},
  eprinttype    = {arXiv},
  eprint       = {2502.07640},
  bibsource    = {dblp computer science bibliography, https://dblp.org}
}

@misc{formalmath,
      title={FormalMATH: Benchmarking Formal Mathematical Reasoning of Large Language Models}, 
      author={Zhouliang Yu and Ruotian Peng and Keyi Ding and Yizhe Li and Zhongyuan Peng and Minghao Liu and Yifan Zhang and Zheng Yuan and Huajian Xin and Wenhao Huang and Yandong Wen and Ge Zhang and Weiyang Liu},
      year={2025},
      eprint={2505.02735},
      archivePrefix={arXiv},
      primaryClass={cs.AI},
      url={https://arxiv.org/abs/2505.02735}, 
}

@misc{illusionthinking,
      title={The Illusion of Thinking: Understanding the Strengths and Limitations of Reasoning Models via the Lens of Problem Complexity}, 
      author={Parshin Shojaee and Iman Mirzadeh and Keivan Alizadeh and Maxwell Horton and Samy Bengio and Mehrdad Farajtabar},
      year={2025},
      eprint={2506.06941},
      archivePrefix={arXiv},
      primaryClass={cs.AI},
      url={https://arxiv.org/abs/2506.06941}, 
}

@inproceedings{llmjudgebeyondaccuracy,
  author       = {Shijie Xia and
                  Xuefeng Li and
                  Yixin Liu and
                  Tongshuang Wu and
                  Pengfei Liu},
  editor       = {Toby Walsh and
                  Julie Shah and
                  Zico Kolter},
  title        = {Evaluating Mathematical Reasoning Beyond Accuracy},
  booktitle    = {AAAI-25, Sponsored by the Association for the Advancement of Artificial
                  Intelligence, February 25 - March 4, 2025, Philadelphia, PA, {USA}},
  pages        = {27723--27730},
  publisher    = {{AAAI} Press},
  year         = {2025},
  url          = {https://doi.org/10.1609/aaai.v39i26.34987},
  doi          = {10.1609/AAAI.V39I26.34987},
  bibsource    = {dblp computer science bibliography, https://dblp.org}
}

@article{beyondaccuracysurvey,
  author       = {Philipp Mondorf and
                  Barbara Plank},
  title        = {Beyond Accuracy: Evaluating the Reasoning Behavior of Large Language
                  Models - {A} Survey},
  journal      = {CoRR},
  volume       = {abs/2404.01869},
  year         = {2024},
  url          = {https://doi.org/10.48550/arXiv.2404.01869},
  doi          = {10.48550/ARXIV.2404.01869},
  eprinttype    = {arXiv},
  eprint       = {2404.01869},
  bibsource    = {dblp computer science bibliography, https://dblp.org}
}

@book{isabelle,
  title={Isabelle/HOL: a proof assistant for higher-order logic},
  author={Nipkow, Tobias and Paulson, Lawrence C and Wenzel, Markus},
  volume={2283},
  year={2002},
  publisher={Springer Science \& Business Media}
}

@inproceedings{favourown,
  author       = {Arjun Panickssery and
                  Samuel R. Bowman and
                  Shi Feng},
  editor       = {Amir Globersons and
                  Lester Mackey and
                  Danielle Belgrave and
                  Angela Fan and
                  Ulrich Paquet and
                  Jakub M. Tomczak and
                  Cheng Zhang},
  title        = {{LLM} Evaluators Recognize and Favor Their Own Generations},
  booktitle    = {Advances in Neural Information Processing Systems 38: Annual Conference
                  on Neural Information Processing Systems 2024, NeurIPS 2024, Vancouver,
                  BC, Canada, December 10 - 15, 2024},
  year         = {2024},
  url          = {http://papers.nips.cc/paper\_files/paper/2024/hash/7f1f0218e45f5414c79c0679633e47bc-Abstract-Conference.html},
  bibsource    = {dblp computer science bibliography, https://dblp.org}
}

@inproceedings{polyrating,
  author       = {Jasper Dekoninck and
                  Maximilian Baader and
                  Martin T. Vechev},
  title        = {Polyrating: {A} Cost-Effective and Bias-Aware Rating System for {LLM}
                  Evaluation},
  booktitle    = {The Thirteenth International Conference on Learning Representations,
                  {ICLR} 2025, Singapore, April 24-28, 2025},
  publisher    = {OpenReview.net},
  year         = {2025},
  url          = {https://openreview.net/forum?id=URPwT55i6O},
  bibsource    = {dblp computer science bibliography, https://dblp.org}
}

@misc{goedelproverv2,
      title={Goedel-Prover-V2: Scaling Formal Theorem Proving with Scaffolded Data Synthesis and Self-Correction}, 
      author={Yong Lin and Shange Tang and Bohan Lyu and Ziran Yang and Jui-Hui Chung and Haoyu Zhao and Lai Jiang and Yihan Geng and Jiawei Ge and Jingruo Sun and Jiayun Wu and Jiri Gesi and Ximing Lu and David Acuna and Kaiyu Yang and Hongzhou Lin and Yejin Choi and Danqi Chen and Sanjeev Arora and Chi Jin},
      year={2025},
      eprint={2508.03613},
      archivePrefix={arXiv},
      primaryClass={cs.LG},
      url={https://arxiv.org/abs/2508.03613}, 
}

@inproceedings{gsm1k,
  author       = {Hugh Zhang and
                  Jeff Da and
                  Dean Lee and
                  Vaughn Robinson and
                  Catherine Wu and
                  William Song and
                  Tiffany Zhao and
                  Pranav Raja and
                  Charlotte Zhuang and
                  Dylan Slack and
                  Qin Lyu and
                  Sean Hendryx and
                  Russell Kaplan and
                  Michele Lunati and
                  Summer Yue},
  editor       = {Amir Globersons and
                  Lester Mackey and
                  Danielle Belgrave and
                  Angela Fan and
                  Ulrich Paquet and
                  Jakub M. Tomczak and
                  Cheng Zhang},
  title        = {A Careful Examination of Large Language Model Performance on Grade
                  School Arithmetic},
  booktitle    = {Advances in Neural Information Processing Systems 38: Annual Conference
                  on Neural Information Processing Systems 2024, NeurIPS 2024, Vancouver,
                  BC, Canada, December 10 - 15, 2024},
  year         = {2024},
  url          = {http://papers.nips.cc/paper\_files/paper/2024/hash/53384f2090c6a5cac952c598fd67992f-Abstract-Datasets\_and\_Benchmarks\_Track.html},
  bibsource    = {dblp computer science bibliography, https://dblp.org}
}

@misc{seedprover,
      title={Seed-Prover: Deep and Broad Reasoning for Automated Theorem Proving}, 
      author={Luoxin Chen and Jinming Gu and Liankai Huang and Wenhao Huang and Zhicheng Jiang and Allan Jie and Xiaoran Jin and Xing Jin and Chenggang Li and Kaijing Ma and Cheng Ren and Jiawei Shen and Wenlei Shi and Tong Sun and He Sun and Jiahui Wang and Siran Wang and Zhihong Wang and Chenrui Wei and Shufa Wei and Yonghui Wu and Yuchen Wu and Yihang Xia and Huajian Xin and Fan Yang and Huaiyuan Ying and Hongyi Yuan and Zheng Yuan and Tianyang Zhan and Chi Zhang and Yue Zhang and Ge Zhang and Tianyun Zhao and Jianqiu Zhao and Yichi Zhou and Thomas Hanwen Zhu},
      year={2025},
      eprint={2507.23726},
      archivePrefix={arXiv},
      primaryClass={cs.AI},
      url={https://arxiv.org/abs/2507.23726}, 
}

@article{zhao2024autograding,
  author       = {Chenyan Zhao and
                  Mariana Silva and
                  Seth Poulsen},
  title        = {Autograding Mathematical Induction Proofs with Natural Language Processing},
  journal      = {CoRR},
  volume       = {abs/2406.10268},
  year         = {2024},
  url          = {https://doi.org/10.48550/arXiv.2406.10268},
  doi          = {10.48550/ARXIV.2406.10268},
  eprinttype    = {arXiv},
  eprint       = {2406.10268},
  bibsource    = {dblp computer science bibliography, https://dblp.org}
}

@inproceedings{zhao2025fewshotgraders,
  author       = {Chenyan Zhao and
                  Mariana Silva and
                  Seth Poulsen},
  editor       = {Alexandra I. Cristea and
                  Erin Walker and
                  Yu Lu and
                  Olga C. Santos and
                  Seiji Isotani},
  title        = {Language Models are Few-Shot Graders},
  booktitle    = {Artificial Intelligence in Education - 26th International Conference,
                  {AIED} 2025, Palermo, Italy, July 22-26, 2025, Proceedings, Part {IV}},
  series       = {Lecture Notes in Computer Science},
  volume       = {15880},
  pages        = {3--16},
  publisher    = {Springer},
  year         = {2025},
  url          = {https://doi.org/10.1007/978-3-031-98459-4\_1},
  doi          = {10.1007/978-3-031-98459-4\_1},
  bibsource    = {dblp computer science bibliography, https://dblp.org}
}

@article{verl,
  title   = {HybridFlow: A Flexible and Efficient RLHF Framework},
  author  = {Guangming Sheng and Chi Zhang and Zilingfeng Ye and Xibin Wu and Wang Zhang and Ru Zhang and Yanghua Peng and Haibin Lin and Chuan Wu},
  year    = {2024},
  journal = {arXiv preprint arXiv: 2409.19256}
}

@inproceedings{chevalier2024science,
  author       = {Alexis Chevalier and
                  Jiayi Geng and
                  Alexander Wettig and
                  Howard Chen and
                  Sebastian Mizera and
                  Toni Annala and
                  Max Jameson Aragon and
                  Arturo Rodr{\'{\i}}guez Fanlo and
                  Simon Frieder and
                  Simon Machado and
                  Akshara Prabhakar and
                  Ellie Thieu and
                  Jiachen T. Wang and
                  Zirui Wang and
                  Xindi Wu and
                  Mengzhou Xia and
                  Wenhan Xia and
                  Jiatong Yu and
                  Junjie Zhu and
                  Zhiyong Jason Ren and
                  Sanjeev Arora and
                  Danqi Chen},
  title        = {Language Models as Science Tutors},
  booktitle    = {Forty-first International Conference on Machine Learning, {ICML} 2024,
                  Vienna, Austria, July 21-27, 2024},
  publisher    = {OpenReview.net},
  year         = {2024},
  url          = {https://openreview.net/forum?id=WFyolnFZOR},
  bibsource    = {dblp computer science bibliography, https://dblp.org}
}
